\documentclass[10pt,compsoc,final]{IEEEtran}
\usepackage{graphicx}
\usepackage{amsmath}
\usepackage{amssymb}
\usepackage{booktabs}
\usepackage{array, siunitx, booktabs}
\usepackage[switch]{lineno}
\usepackage{cases}
\usepackage[table,xcdraw]{xcolor}
\usepackage{soul}
\usepackage{wasysym}

\usepackage{color}
\usepackage{tabularray}

\usepackage{xcolor}
\usepackage{tikz}

\definecolor{good_col}{RGB}{68,1,84}
\definecolor{bad_col}{RGB}{253,231,57}

\newcommand{\priority}[1]{
	\tikz[node distance=1mm baseline]
	{
		\draw[fill=bad_col, draw=none] (0,0) circle (.07);
		\draw[fill=good_col, draw=none]  (0,0) -- (.07,0) arc[start angle=0, delta angle=#1,radius=.07cm] -- (0,0);
	}
}

\definecolor{Changes}{RGB}{0, 0, 0}

\usepackage[pagebackref,breaklinks,colorlinks]{hyperref}
\usepackage{setspace}

\ifCLASSOPTIONcompsoc
\usepackage[nocompress]{cite}
\else
\usepackage{cite}
\fi

\graphicspath{{./imgs_jpg/}}
\newcommand{\PAR}[1]{\left( #1 \right)}
\newcommand{\ournerf}{Season-NeRF}

\begin{document}
	%\linenumbers
	%\doublespacing
	\title{Incorporating Season and Solar Specificity into Renderings made by a NeRF Architecture using Satellite Images}

	%\author{anonymous authors}
	\author{Michael Gableman and Avinash Kak}
	%Authors: Michael Gableman and Avi Kak
	
	%		\IEEEcompsocitemizethanks{\IEEEcompsocthanksitem M. Gableman is a student at Purdue University\protect\\
		%			% note need leading \protect in front of \\ to get a newline within \thanks as
		%			% \\ is fragile and will error could use \hfil\break instead.
		%			E-mail: mgablema@purdue.edu}
	%		\thanks{Manuscript received April 19, 2005; rev ised August 26, 2015.}}
%

%\markboth{IEEE TRANSACTIONS ON PATTERN ANALYSIS AND MACHINE INTELLIGENCE,~Vol.~14, No.~8, August~2015}%
%{Shell \MakeLowercase{\textit{et al.}}: Bare Demo of IEEEtran.cls for Computer Society Journals}
%\markboth{IEEE TRANSACTIONS ON PATTERN ANALYSIS AND MACHINE INTELLIGENCE}{}

\IEEEtitleabstractindextext{%
	%%%%%%%%% ABSTRACT
	\begin{abstract}
		As a result of Shadow NeRF and Sat-NeRF, it is possible to take the solar angle into account in a NeRF-based framework for rendering a scene from a novel viewpoint using satellite images for training.  Our work extends those
		contributions and shows how one can make the renderings season-specific.  
		Our main challenge was creating a Neural Radiance Field (NeRF) that could render seasonal features independently of viewing angle and solar angle while still being able to render shadows.
		We teach our network to render seasonal features by introducing one
		more input variable --- time of the year.
		However, the small training datasets typical of satellite imagery can introduce ambiguities in cases where shadows are present in the same location for every image of a particular season.
		We add additional terms to the loss function to discourage the network from using seasonal features for accounting for shadows.
		We show the performance
		of our network on eight Areas of Interest containing images captured by the Maxar WorldView-3 satellite.
		This evaluation includes tests measuring
		the ability of our framework to accurately render novel views, generate height maps, predict shadows, and specify seasonal features independently from shadows.
		Our ablation studies justify the choices made for network design parameters.
	\end{abstract}
	% Note that keywords are not normally used for peer review papers.
	\begin{IEEEkeywords}
		NeRF, Satellite Imagery, Image-based rendering, Time-varying imagery
\end{IEEEkeywords}}

\maketitle

\IEEEdisplaynontitleabstractindextext
\IEEEpeerreviewmaketitle

\begin{figure}[!t]
	\centering
	\includegraphics[width=1.0\linewidth]{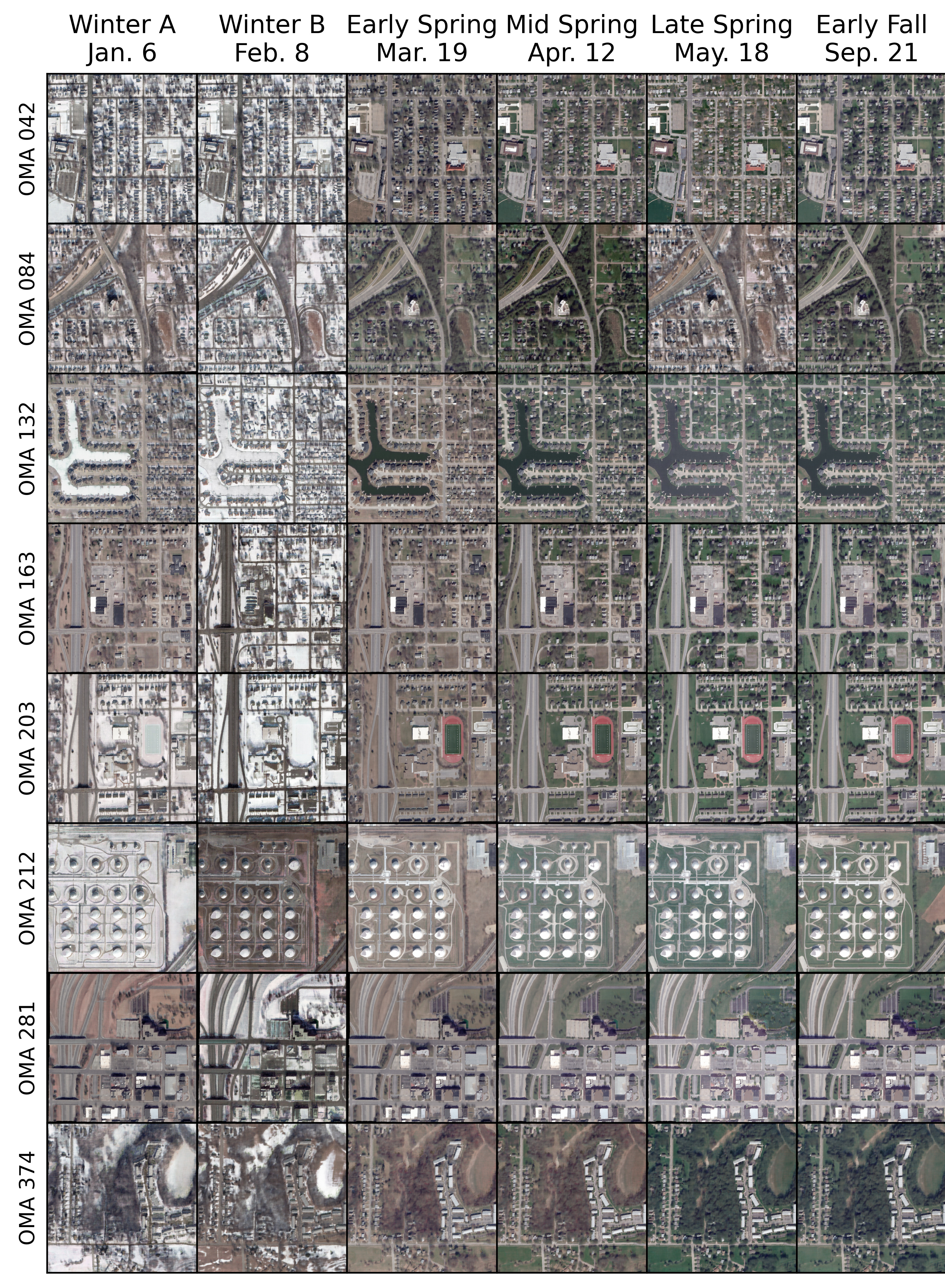}
	\caption{\label{Fig:Summary_Img} Results of generating images
		on eight regions.  Each view is generated at 26 degrees off-nadir with a 45-degree azimuth angle.  The solar angle is at
		a 50 degrees elevation and 135-degree azimuth.  Images in
		the same columns have the same time input.}
\end{figure}

\IEEEraisesectionheading{\section{Introduction}\label{Sec:Introduction}}
\IEEEPARstart{O}{ur} goal in this paper is to extend the NeRF
formalism to render season-specific images with correct shadows from multidate
and multiview satellite images.  What we seek to accomplish is
exemplified by the results shown in Fig. \ref{Fig:Summary_Img} in
which the columns represent different seasons, and the eight rows are the eight different Areas of Interest (AOIs) we tested from the 2019 IEEE
GRSS dataset \cite{Sat_data_2}.
These rendered images are from viewpoints, solar angles, and times of the year not used for training the NeRF.
The examples shown in Fig. \ref{Fig:Summary_Img} use the time of the year to alter the seasonal features while keeping the viewing and solar angles constant.
\textcolor{Changes}{In addition, we provide an animated example of changing viewpoints with fixed time of year and solar angle on our GitHub page\footnote{https://github.com/EnterpriseCV-6/Season-NeRF.git}, where our code is also available.}
Furthermore, the figure shows that the seasonal features in the rendered image correlate with the time of the year given as input.

The original NeRF formulation, presented by Mildenhall et al. \cite{NeRF}, did not have mechanisms to account for changes in an image unrelated to changes in viewing angle.
This restriction on NeRFs was relaxed by Martin-Brualla et al. \cite{WNeRF} and by Derksen and Izzo \cite{SNeRF}.  
Both works accounted for non-viewing angle image changes by introducing additional input into the NeRF.
NeRF in the Wild (NeRF-W) \cite{WNeRF} accomplished this by using two learnable embeddings for each training image.
One embedding captured the {\em appearance} of the image, in essence, the information that could explain alterations in the manifestation of objects in the scene.
The other embedding captured the {\em transient} portions of the image.  These portions of the image contain moving objects that should not appear in a rendered image.
To avoid a trivial solution where every point in every image is deemed transient, NeRF-W uses the Bayesian learning framework of Kendall et al. \cite{Bayesian_WNeRF_Frame}.
More details on NeRF-W are provided in Section \ref{Sec:Related_Works}.

Derksen and Izzo \cite{SNeRF} noted that a large amount of image variation in satellite images is caused by changing light conditions and factored this observation directly into the NeRF model.
They used the twin concepts of {\em solar
	visibility} and {\em sky color} as two additional outputs of their
NeRF framework, known as Shadow NeRF (S-NeRF).
The solar visibility computes the amount of direct sunlight at each point in the model, and the sky color accounts for global light conditions and models the background illumination.
To compute these additional outputs, S-NeRF takes a solar angle as input to the network.
S-NeRF limits the influence of the solar angle by only allowing it to influence the rendering in a manner consistent with how light interacts with surfaces.
More details on how S-NeRF uses solar visibility and sky color are provided in Section \ref{Sec:Related_Works}.
In addition to outputting solar visibility and sky color, S-NeRF replaces the computation of color with the computation of albedo color.

What has motivated the work reported in NeRF-W \cite{WNeRF} and
S-NeRF \cite{SNeRF} is exactly our motivation as well.
Satellite images of
the same scene may be captured months apart.
As a result, two
satellite images may exhibit different seasonal features and shadows while containing transient objects.
We seek to create a NeRF framework capable of accounting for the seasonal features and shadows while simultaneously discarding portions of the training images containing transient objects.
Transient objects, such as cars moving along a road, appear in only a single image and should \textcolor{Changes}{not be allowed to} influence the final rendering.
Seasonal features manifest themselves differently during different times of the year.
For example, images may contain large swathes of
green foliage from late spring to early fall.
However, it is common
to see snow and brown foliage from late fall to early spring.
Shadows are caused by objects occluding light rays from the sun to a world point.

The framework we present in this paper, which we call \ournerf{}, computes the seasonally adjusted albedo, density, sky color, and solar visibility as a function of position within the model, solar angle, and time of the year.
The solar angle and the time of the year are provided in the metadata associated with the satellite image.
We wish to ensure that any changes in the time of the year only change the seasonal features, and changes to the solar angle only change shadows.
Without limitations, the time of the year and solar angle can manipulate the rendering in ways unassociated with their respective features.
To constrain the use of the additional inputs, we only allow the time of the year to alter the value of the seasonally adjusted albedo, and we only allow the solar angle to alter the values of the sky color and solar visibility.
We also limit how much the time of the year can alter the seasonally adjusted albedo and how solar visibility and sky color interact with seasonally adjusted albedo.
More details on how the time of the year influences the albedo color are provided in Section \ref{Sec:Seasonal_Adjustment}, and we describe the computation of shadows in Section \ref{Sec:Shadow_adjust}.

Accurate placement of shadows within an image and correct novel view rendering require an accurate density estimation.
To encourage accurate density estimation, we provide a height map to \ournerf{} during the early stages of training.
Our use of structural information is similar to the technique used by Deng et al. \cite{Depth_Supervised_NeRF}.
In a depth-supervised NeRF, traditional computer vision methods extract 3D world points.
These points are used to provide depth supervision during training.
We use this height map to guide the density computation in the early stages of training.
{\em In the later stages, we stop using the height map} as the map is noisy, and the partially trained \ournerf{} can provide more accurate structural estimations.
More details on our use of a height map are provided in Section \ref{Sec:MultiPhase_Training}.

In addition to solar and seasonal variation in images, we also account for transient objects.
Transient objects are moving objects that appear in single images and should be excluded from the final model.
NeRF-W uses learned image-specific embeddings to compute the network's uncertainty of the predicted color.
The network learns to assign regions with transient objects a higher uncertainty.
We propose an alternative that removes the need to learn image-specific embeddings.
The loss proposed by Barron \cite{Robust_Loss}
allows the network to reduce the influence of transient objects automatically.
This loss allows us to use a simpler architecture, removes the need to compute and account for an uncertainty field, and simplifies the rendering process during training.

When evaluating \ournerf{}'s performance, we show that \ournerf{} can render high-quality images, stably render
seasons, and correctly specify seasons and shadows.  Due to the limited data, we can only withhold four images from each region
for testing.
Of the four images withheld for testing, we ensure that three of
these images span unique seasons.
The fourth image is selected to ensure that the set of testing images in each AOI spans a reasonable range of solar and viewing angles.
The quality of the rendering is determined by comparing the
testing images with rendered images at the same viewing and solar
angle.  In addition to image similarity, we generate a height map of
the region and compare that to the lidar data associated with the AOI.
The accuracy of our estimated shadow masks is computed by comparing the estimated mask to
the exact mask \ournerf{} computes.
Stably rendering a season means that the seasonal features should only change due to changing the time of the year, not because of changes in the solar or the viewing angles.
To measure this, we render images across a wide range
of viewing angles and solar angles.
The extent of the change introduced by altering the viewing and solar angle is measured against the extent of the change introduced
by altering viewing time.
Correctly specify seasons (i.e., getting a
winter image for a time input during the winter or a spring image for a time input during the spring) is confirmed by rendering images across the entire year and
visually confirming the correct seasonal characteristics.

The remainder of this paper is organized as follows,
Section \ref{Sec:Related_Works} discusses works related to our approach.
In Section \ref{Sec:Process}, we provide details on the specifics of our
approach.  Section \ref{Sec:Results} contains the results of \ournerf{}
when applied to eight different regions and an ablation study.
We break our analysis into three subsections.
In Section \ref{Sec:Novel_View_Results}, we show how \ournerf{} can render novel views.
Then, in Section \ref{Sec:HM_Results}, we show how \ournerf{} can generate a height map and predict shadows, and in Section \ref{Sec:Season_Specific_and_Stable_Results}, we show the seasonal stability and specificity of \ournerf{}.
In addition, within each subsection, we show how different components contribute to the accuracy of the final model.
Our results and conclusions are summarized in
Sec. \ref{Sec:Conclusion}.

\section{Related Works}
\label{Sec:Related_Works}
Neural Radiance Fields (NeRFs) have become a popular tool for novel view reconstruction since they were used by Mildenhall et al. \cite{NeRF}.
Since this work, NeRFs have been used with success in many situations.
Work with NeRFs generally falls into three categories, improved novel view generation \cite{NeRV, SNeRF, Sat_NeRF, MipNeRF, NeRFMVS, RNeRF, WNeRF, NeRF++, Nerfies}, reduced reliance on camera parameters \cite{GNeRF, BNeRF}, and faster model generation \cite{MVSNeRF, DietNeRF, pixelNeRF, Depth_Supervised_NeRF}.
While all these areas are relevant to understanding NeRFs, we are primarily interested in improved novel view generation, specifically for satellite images.

A NeRF computes a density, $\rho$\footnote{In \cite{NeRF}, $\sigma$ is used for density instead of $\rho$, however in this paper $\sigma$ is used to indicate the sigmoid non-linearity function.}, and a color, $C$, for every point within a 3-dimensional world space.
This computation is carried out by a neural network trained on pixel-ray pairs.
Every pixel has a known color, and its corresponding ray is the preimage of the pixel in the world space.
The NeRF uses a ray to render a pixel's color by considering the color and density of every point along the ray.
This computation requires integration along the ray.
However, such an approach is not practical, so the quadrature rule is used to approximate the integral along the ray.
The rendered color of a pixel with a preimage ray $\chi$ containing points $\vec{X}_0, \vec{X}_1, \cdots, \vec{X}_n$ is
\begin{equation}
	\label{Eq:NeRF_Equation}
	Col\PAR{\chi} = \sum_{\vec{X}_i\in\chi} C\PAR{\vec{X}_i}P_S\PAR{\chi, \vec{X}_i},
\end{equation}
where $P_S$ is the probability that $\vec{X}_i$ is a surface point for ray $\chi$, $C$ is the color output by the NeRF at $\vec{X}_i$, and $Col$ is the rendered color of the pixel.
A point on a ray is the surface point for the ray if the point exists (that is, the space is occupied) and the point is visible along the ray.
Furthermore, a point on a ray is visible along the ray if no prior points along the ray exist.
Thus,
\begin{equation}
	\label{Eq:PSurf}
	P_S\PAR{\chi, \vec{X}_i} = P_E\PAR{\vec{X}_i} P_V\PAR{\vec{X}_i, \chi},
\end{equation}
\begin{equation}
	\label{Eq:PE}
	P_E\PAR{\vec{X}_i} = 1-e^{-\rho\PAR{\vec{X}_i} \delta_{\vec{X}_i}},
\end{equation}
and
\begin{equation}
	\label{Eq:PV}
	P_V\PAR{\vec{X}_i, \chi} = e^{-\sum\limits_{j=0}^{i-1} \rho\PAR{\vec{X}_j} \delta_{\vec{X}_j}}
\end{equation}
where $\delta_{\vec{X}_i}$ is the distance between the points $\vec{X}_i$ and $\vec{X}_{i+1}$ and $\rho$ is the density at $\vec{X}_i$\footnote{The symbols $P_S$ and $P_V$ are analogous to $w_i$, $T_i$ from \cite{NeRF}, however as $\vec{w}$ is used for solar angle and $T$ is associated with time in this work, we introduce new notation.}.
Initial work by Mildenhall et al. \cite{NeRF} suggested using a weighted stratified sampling scheme for determining points along a ray; however, Mari et al. \cite{Sat_NeRF} use a uniform stratified sampling scheme.
We are not aware of any work comparing the two schemes and use uniform stratified sampling throughout our work.

The work of Derksen and Izzo \cite{SNeRF} provided a means for a NeRF to account for shadows in a scene.
They accounted for shadows by expanding NeRF's outputs to include solar visibility ($S_{vis}$) and sky color ($\text{sky}$), which were computed by considering the solar angle.
Their variant of NeRF is known as S-NeRF.
Rather than outputting color, S-NeRF outputs albedo color ($A$).
S-NeRF also modifies the rendering process to incorporate the new output terms.
The color portion $C$ from Equation \ref{Eq:NeRF_Equation} is replaced with
\begin{equation}
	\begin{split}
		\label{SNeRF_Color}
		C\PAR{\vec{X}, \vec{w}} = &\left( S_{vis}\PAR{\vec{X}, \vec{w}} \right. \left. + \PAR{1-S_{vis}\PAR{\vec{X}, \vec{w}}} \text{sky}\PAR{\vec{w}} \right) \\
		& *A\PAR{\vec{X}},
	\end{split}
\end{equation}
where $\vec{w}$ is the solar angle.
For S-NeRF to correctly render shadows, the model trains with two types of rays, {\em image rays} and {\em solar rays}.
Image rays are associated with a pixel from the training data and a solar angle.
They are used to ensure the network accurately renders scenes.
Solar rays are not associated with the training data, and the solar angle associated with a solar ray is negative one times the direction of the solar ray.
Solar rays are used to ensure that the solar visibility output by the network is consistent with the density output of the network.
By adding these modifications, S-NeRF can account for shadows and changing solar conditions during training and evaluation.
\textcolor{Changes}{Allowing the sky color to vary with the solar angle allows S-NeRF to account for changes in the image's appearance caused by different atmospheric effects.}
Furthermore, S-NeRF uses sinusoidal representation networks (SIRENs) as described in Sitzmann et al. \cite{SIREN}.

Another method for accounting for changes in appearance was suggested in Martin-Brualla et al. \cite{WNeRF}.
They created a modification to NeRF known as NeRF-W.
In NeRF-W, each training image is associated with a learned embedding.
The network uses this embedding to output a density and color adjustment for each image, which is used during training to account for features unique to the training image.
The density and color adjustments are ignored during the evaluation.
Furthermore, NeRF-W outputs an uncertainty estimate $\beta$ in addition to the color and density adjustments.
Drawing from the work of \cite{Multi_task_learning}, this uncertainty changes the loss from Mean Squared Error (MSE) loss to 
\begin{equation}
	\label{eq_w_nerf}
	L = \frac{\left|\left| C_{GT} - C_{Est}\right|\right|_2^2}{2\beta_\Sigma^2} + \frac{\log\PAR{\beta_\Sigma^2}}{2}
\end{equation}
where
\begin{equation}
	\beta_\Sigma = \sum\limits_{\vec{X}_i \in \chi} P_{S,T}\PAR{\chi, \vec{X}_i} \beta_i
\end{equation}
where $C_{GT}$ is the ground truth color, $C_{Est}$ is the estimated color, \textcolor{Changes}{and $P_{S,T}$ is the transient probability that $\vec{X}_i$ is the surface for $\chi$}.
The model learns to assign transient objects a higher uncertainty than non-transient objects.
This results in the model lowering the contribution of the transient objects to training loss.

In addition to incorporating ideas from S-NeRF and NeRF-W, Mari et al. \cite{Sat_NeRF} used bundle adjustment to refine the parameters modeling the relationship between pixels and rays.
Their model, known as Sat-NeRF, directly utilizes the Rational Polynomial Coefficient (RPC) approximation of the satellite camera model when determining rays associated with pixels.
The RPC model is described in Grodecki, and Dial's work \cite{RPC_BA}.
The RPC model for satellites is a popular model used to approximate the physical satellite camera model.
Sat-NeRF also uses the 3D points extracted during bundle adjustment to apply the depth supervision technique described in Deng et al. \cite{Depth_Supervised_NeRF}.
Like NeRF-W, Sat-NeRF learns embeddings for each image and outputs uncertainty to account for transient objects.
However, they forgo the image-specific density and color modifications in NeRF-W.

As an alternative to the method used in \cite{WNeRF} and \cite{Sat_NeRF} to account for transient objects, we use the loss function described by Barron in \cite{Robust_Loss}.
This loss function is defined as
\begin{equation}
	\label{Eq_Gen_Loss}
	L_{\alpha, c}\PAR{x} = -\log\PAR{\frac{1}{cZ\PAR{\alpha}}\exp\PAR{-f\PAR{x,\alpha,c}}},
\end{equation}
\begin{equation}
	\label{Eq_Gen_loss_no_log}
	f\PAR{x,\alpha, c}
	=
	\frac{\left| \alpha - 2 \right|}{\alpha}
	\PAR{\PAR{\frac{\PAR{x/c}^2}{\left|\alpha-2\right|}+1}^{\alpha/2}-1},
\end{equation}
and
\begin{equation}
    \label{EQ_Merged}
	Z\PAR{\alpha} = \int\limits_{-\infty}^{\infty} \exp\PAR{-f\PAR{x,\alpha,1}}dx.
\end{equation}
In the above equations, $x$ refers to the difference between the predicted and actual values, and $\alpha$ and $c$ are both learnable parameters.
In Equation \ref{Eq_Gen_loss_no_log}, if $\left|x\right| < c$, the function behaves like the MSE loss.
For values of $\left|x\right| > c$, Equation \ref{Eq_Gen_loss_no_log}'s behavior is determined by $\alpha$.
As $\alpha$ approaches two, the loss behaves like the MSE loss, and as $\alpha$ approaches zero, the loss behaves like the Cauchy loss.
Thus the larger the value of $\alpha$, the larger the contribution to the gradient for errors when $\left|x\right| > c$.
Theoretically, $\alpha$ can be any real number, and $c$ can be any positive number; however, in practice, $\alpha$ is forced to remain in the range $\PAR{0,2}$ and $c$ remains in the range $\PAR{0,1}$.
Note that Equation \ref{Eq_Gen_Loss} is the negative log-likelihood of a probability density function created from the normalized form of Equation \ref{Eq_Gen_loss_no_log}.
Using Equation \ref{Eq_Gen_Loss} instead of Equation \ref{Eq_Gen_loss_no_log} prevents the network from converging to a trivial solution by minimizing $\alpha$.

\section{Method}
\label{Sec:Process}
\ournerf{}, like all variants of NeRF, computes the density and color of every point within the world.
\ournerf{} uses the time of the year and the location within the world to determine the color of world points.
These input terms allow our approach to account for seasonal variation within the image set.
\ournerf{} uses solar angle and location within the model to compute solar visibility.
Solar angle alone is used to compute the sky color.
The solar visibility and sky color of points along a ray are used to compute a shadow adjustment term for rendered pixels.
By using the loss function proposed in Barron \cite{Robust_Loss}, we attempt to reduce the influence that transient objects in the training images have on the model.

\subsection{Seasonal Variation Adjustment}
\label{Sec:Seasonal_Adjustment}
To account for the seasonal changes, we alter the computation of the albedo color by adding a seasonal adjustment term.
We refer to the new term as the seasonally adjusted albedo.
The seasonal adjustment term is a function of position within the model and the time of the year ($t$).
The sigmoid non-linearity is not applied to the seasonally adjusted albedo until after the albedo color has been merged with the seasonal adjustment term.
\textcolor{Changes}{Applying the non-linearity after the combination of the albedo color and seasonal adjustment allows the combined term to take on any value before the non-linearity.
		After the non-linearity has been applied the color will fall in the valid color range of $\left\lbrack 0,1\right\rbrack$.}
Thus the seasonally adjusted albedo is expressed as
\begin{equation}
	\label{EQ:Col_Adj}
	A_t\PAR{\vec{X}, t} = \sigma\PAR{A^*\PAR{\vec{X}} + T_a\PAR{\vec{X}, t}}
\end{equation}
where $A^*$ is the albedo color before the sigmoid non-linearity\textcolor{Changes}{, $\sigma$,} is applied and $T_a$ is the seasonal adjustment term.

To compute the seasonal adjustment term, we use two intermediate
terms, which we refer to as temporal class and temporal adjustment,
$T_C\PAR{t}$ and $T_A\PAR{\vec{X}}$.
Temporal class is dependent on the time of the year, and temporal adjustment is
dependent on the position within the model.
The temporal class is a $N \times 1$ matrix, where $N$ is a
hyper-parameter describing the number of different seasonal classes the model can render.
We use softmax to ensure the values in $T_C$ form a discrete probability distribution over the seasonal classes.
The temporal adjustment is a $C \times N$ matrix, where $C$ is the number of output channels in the rendered image.
The seasonal adjustment is
\begin{equation}
	T_a\PAR{\vec{X}, t} = T_A\PAR{\vec{X}} \times T_C\PAR{t}.
\end{equation}
Intuitively, the $i$th position of $T_C\PAR{t}$ represents the probability that time $t$ corresponds to season $i$, and the columns of $T_A\PAR{\vec{X}}$ represent the seasonal adjustments for each seasonal class.
This results in $T_a$ being the expected seasonal adjustment given the probability of each seasonal class $T_C$.
If $N$ is too large (that is, the model can generate too many possible seasons),
shadow effects will be absorbed into the seasonal adjustment.
If $N$ is too small, the model lacks the descriptive capability to render
every season.
We let $N=4$ for our tests, resulting in a good balance between solar and temporal changes.

Our model only allows $t$ to change the value of the seasonally adjusted albedo, leaving density, solar visibility, and sky color constant with respect to $t$.
\textcolor{Changes}{The limitation that a region's density cannot change over time is an unrealistic assumption, as changes in vegetation can cause the density to change. However, we do not allow our model's density to vary over time for two reasons.
			First, limiting the computation in this way prevents the model from using $t$ to change the density within the world or shadows within rendered images to account for purely visual changes of objects with fixed densities, such as buildings.
			Second, each region contains a single ground truth height map, making accurate evaluation of the quality of changing height maps caused by changing densities difficult.}
The seasonally adjusted rendered color is
\begin{equation}
	Col_t\PAR{\chi, t} = \sum\limits_{\vec{X}_i \in \chi} A_t\PAR{\vec{X}_i, t} P_S\PAR{\chi, \vec{X}_i}
\end{equation}
and is used instead of Equation \ref{Eq:NeRF_Equation}.

To compute the seasonally adjusted rendered color, we must input the time of the year into the network.
Rather than directly sending the day and month to the network, we encode the time as
\begin{equation}
	\label{Eq_time_encode}
	t = \PAR{\cos\PAR{2\pi t_p}, \sin\PAR{2\pi t_p}}
\end{equation} 
where $t_p$ is the fraction of the year completed.
Given two times of the year,
$t_0$ and $t_1 = t_0 + \delta$, the L2 distance between the two encodings is
$2\left|\sin\PAR{\pi \delta}\right|$.
The L2 distance achieves its maximum value at
$\delta=1/2$, corresponding to the day at the opposite end of the year.
This approach also ensures that days at the beginning and end of the year have similar encodings.

\subsection{Shadow Adjustment}
\label{Sec:Shadow_adjust}
Like S-NeRF, \ournerf{} outputs $S_{vis}\PAR{\vec{X}, \vec{w}}$ and $\text{sky}\PAR{\vec{w}}$ however we do not apply these terms directly to the seasonally adjusted albedo.
Instead, we use these terms to compute a shadow mask, which measures the probability that a rendered pixel is \textcolor{Changes}{not} in shadow.
The equation that describes the shadow mask is
\begin{equation}
	\label{Eq:Shadow_Mask}
	M\PAR{\chi, \vec{w}} = \sigma\PAR{\kappa\PAR{\mu+\sum_{\vec{X}_i \in \chi} P_S\PAR{\chi, \vec{X}_i} S_{vis}\PAR{\vec{X}_i, \vec{w}}}},
\end{equation}
where $\vec{w}$ is the solar angle, $S_{vis}$ is the solar visibility from the network, and $\sigma$ is the sigmoid activation function.
\textcolor{Changes}{Note that $M\PAR{\chi, \vec{w}}$ is the probability that the the surface of $\chi$ is visible from the sun.
Therefore, $1-M\PAR{\chi, \vec{w}}$ is the probability that the surface of $\chi$ is in shadow.}
The hyperparameter $\kappa$ controls the rapidity of the transition between non-shadow and shadow.
The hyperparameter $\mu$ controls the threshold where this transition occurs.
We let $\kappa = 30$ and $\mu = -.2$, as we found these perform well for shadow generation.

Given a shadow mask and a seasonally adjusted rendered color, the shadow-and-seasonally adjusted rendered color is
\begin{equation}
	\label{Eq:Shadow_Modded_Color}
	\begin{split}
		Col_{SA}&\PAR{\chi, \vec{w}, t} = Col_t\PAR{\chi, t}*\\
		& \PAR{ M\PAR{\chi, \vec{w}} + \PAR{1-M\PAR{\chi, \vec{w}}}\text{sky}\PAR{\vec{w}}}.
	\end{split}
\end{equation}
Equation \ref{Eq:Shadow_Modded_Color} assumes that $Col_t\PAR{\chi, t}$ is the color of the pixel in direct sunlight and $\text{sky}\PAR{\vec{w}} Col_t\PAR{\chi, t}$ is the color of the pixel when the pixel is in shadow.
The value of $M\PAR{\chi, \vec{w}}$ is the probability that the rendered pixel is in shadow.
Other sources than direct light from the sun may still illuminate regions in shadow.
As such, we allow $\left|\left|\text{sky}\PAR{\vec{w}}\right|\right|_2$ to be greater than zero to ensure regions in shadow are partially lit.

\subsection{The Loss Function and Network Architecture}
We pass solar and image rays through the network during training.
As with S-NeRF, our loss function behaves differently for solar and image rays.
The loss function associated with image rays consists of three terms.
The first term applies Barron's loss to the absolute difference in the RGB color of the rendered pixel and the ground truth pixel.
We use Barron's loss to reduce the effect of transient objects on the training process.
The second loss term encourages at least one channel in the seasonally adjusted rendered color to be above a user-defined threshold.
By encouraging the model to have at least one large channel in $Col_t\PAR{\chi, t}$, we avoid using the seasonally adjusted albedo in place of shadows to describe dark regions.
The second loss term is
\begin{equation}{L_A\PAR{\vec{C}} = \frac{1}{n}\sum\limits_{i=1}^{3}}
	\PAR{1 - \frac{\min\PAR{\vec{C}_i, \mathbb{A}}}{\mathbb{A}}}^2
\end{equation}
where $\mathbb{A}$ is a hyperparmeter such that $\mathbb{A} \in \left(0, 1\right\rbrack$.
We use $\mathbb{A} = 0.2$ as most non-shadow regions of our images have at least one channel above this value.
Even with this loss term, our network tended to ignore shadows by setting the sky color to one.
To avoid this, we added a third loss function
\begin{subnumcases}{\label{Eq_Sky_loss}
		L_{sky}\PAR{\text{sky}} = \sum\limits_{i=1}^{3}}
	0 & $\text{sky}_i \leq \mathbb{S}$ \\
	\PAR{\frac{1}{\mathbb{S}}*\text{sky}_i-1}^2 & $\text{sky}_i > \mathbb{S}$,
\end{subnumcases}
where $\mathbb{S}$ is a hyperparmeter such that $\mathbb{S} \in \left(0, 1\right\rbrack$.
We use $\mathbb{S} = 0.5$ because this will encourage areas in shadow to reduce the seasonally adjusted albedo color by at least 50\% while still allowing some indirect light to illuminate the region.
Given an image ray $\PAR{\chi, \vec{w}, t, \vec{C}_{GT}}$, the total loss associated with that ray is
\begin{equation}
	\begin{split}
		L_{\text{IR}} =  &L_{\alpha, c}\PAR{\vec{C}_{GT} - Col_{SA}\PAR{\chi, \vec{w}, t}}\\
		& + L_A\PAR{Col_t\PAR{\chi, t}} + L_{sky}\PAR{\text{sky}\PAR{\vec{w}}}.
	\end{split}
\end{equation}

We shall now describe the training loss associated with solar rays.
The loss function associated with solar rays consists of two terms.
The first term minimizes the MSE between the estimated solar visibility, $S_{vis}\PAR{\vec{w}, \chi}$ and the exact solar visibility computed by Equation \ref{Eq:PV}.
The second term is the loss described in Equation \ref{Eq_Sky_loss}.
Applying Equation \ref{Eq_Sky_loss} to solar rays ensures the sky color is appropriate, even if the input solar angle is far from the solar angles from the training set.
The total loss for a solar ray is
\begin{equation}
	\begin{split}
		L_{\text{SR}} = & L_{sky}\PAR{\text{sky}\PAR{\vec{w}}} + \\
		&\frac{1}{K}\sum\limits_{j=1}^{K} \PAR{S_{vis}\PAR{\vec{X}_{j}, \vec{w}} - P_V\PAR{\vec{X}_{j}, \chi}}^2.
	\end{split}
\end{equation}
Unlike S-NeRF, we do not use an absorption term for solar ray loss.
The accuracy of \ournerf{}'s shadow mask is limited by the accuracy of its
density computation because the training process uses the density output of the network to determine the solar visibility.

\begin{figure}[!t]
	\centering
	\includegraphics[width=1.0\linewidth]{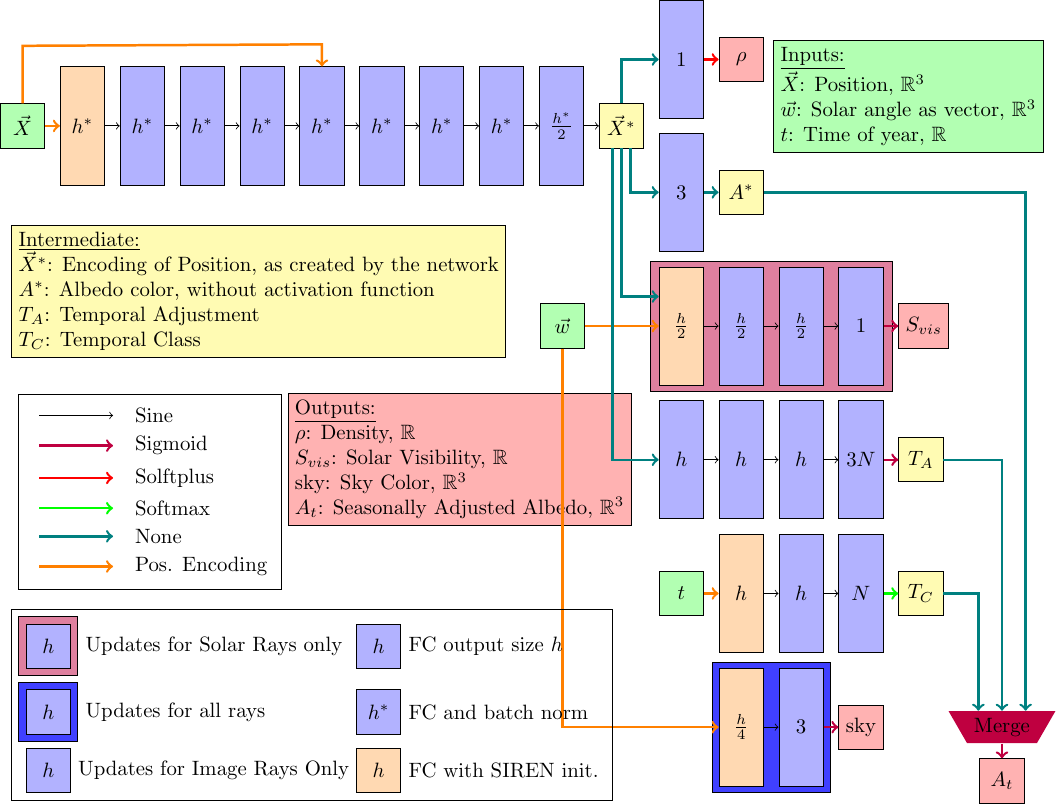}
	\caption{\label{Fig:network} \ournerf{}'s network architecture where, $\vec{X}$ is the input's spatial coordinates, $\vec{w}$ is the solar angle, and $t$ is the time of input.
		The model predicts density, seasonally adjusted albedo, solar visibility, and sky color.
		The Merge trapezoid indicates the effect of Equation \ref{EQ:Col_Adj} and does not contain any learnable parameters.}
\end{figure}

We shall now describe the architecture of \ournerf{}.
\ournerf{} consists of SIREN fully connected layers.
\textcolor{Changes}{Some of these layers include batch normalization, despite batch normalization being non-standard with NeRFs and SIRENs, as we found this gave improved results, as shown in Table \ref{Tab:Batch_Norm}\footnote{
Our tables employ the symbol $\uparrow$ to represent metrics where larger numbers signify superior results and $\downarrow$ for metrics where smaller numbers signify superior results.
In addition, we use \priority{360} and \textbf{bold} to indicate the result with the best score.
The symbol
\priority{0} to indicate the worst score.}.
Table \ref{Tab:Batch_Norm} uses the Structural Similarity Index Measure (SSIM) to measure image quality and Mean Absolute Error (MAE) to measure the quality of the height map.}
Furthermore, the inputs of our network undergo positional encoding, as described in \cite{NeRF}, before being processed.
We provide a network diagram of \ournerf{} in Fig. \ref{Fig:network}.
We set the layer width of the network to 512.
During training, we use 512 image rays and 1024 solar rays per batch, with all rays sampled at 96 points.
When backpropagating the loss for image rays, we freeze weights within the model used only to compute solar visibility.
Also, when backpropagating the loss for solar rays, we freeze weights within the model that are not used exclusively for the computation of sky color or solar visibility.
We use the One Cycle Learning Rate Scheduler described by Smith and Topin \cite{LR3} during training with a maximum learning rate of $1.5\mathrm{e}{-4}$.
We train our network for 50000 steps, which takes approximately 7 hours on an Nvidia GeForce GTX 1080 Ti GPU.
\begin{table*}[!htb]
			\centering
			\caption{\label{Tab:Batch_Norm}Comparison of results with and without batch normalization.}
			\begin{tabular}{|c|c c | c c | }
				\hline
				\text{Region} & \multicolumn{2}{c|}{\text{SSIM, SA $\uparrow$}} & \multicolumn{2}{c|}{\text{MAE $\downarrow$}} \\ \hline
				\text{Case:} & \text{With Batch Norm} & \text{No Batch Norm} & \text{With Batch Norm} & \text{No Batch Norm} \\ \hline
				\text{OMA 042} &  0.562 \priority{0} &  \textbf{0.562} \priority{360} &  \textbf{2.357} \priority{360} &  2.496 \priority{0}  \\ \hline
				\text{OMA 084} &  \textbf{0.529} \priority{360} &  0.514 \priority{0} &  \textbf{3.769} \priority{360} &  4.684 \priority{0}  \\ \hline
				\text{OMA 132} &  0.704 \priority{0} &  \textbf{0.709} \priority{360} &  \textbf{1.279} \priority{360} &  1.524 \priority{0}  \\ \hline
				\text{OMA 163} &  \textbf{0.527} \priority{360} &  0.518 \priority{0} &  \textbf{2.073} \priority{360} &  2.311 \priority{0}  \\ \hline
				\text{OMA 203} &  \textbf{0.696} \priority{360} &  0.691 \priority{0} &  \textbf{1.429} \priority{360} &  2.586 \priority{0}  \\ \hline
				\text{OMA 212} &  \textbf{0.679} \priority{360} &  0.670 \priority{0} &  \textbf{1.102} \priority{360} &  1.212 \priority{0}  \\ \hline
				\text{OMA 281} &  \textbf{0.618} \priority{360} &  0.617 \priority{0} &  \textbf{2.296} \priority{360} &  3.476 \priority{0}  \\ \hline
				\text{OMA 374} &  0.473 \priority{0} &  \textbf{0.474} \priority{360} &  \textbf{4.211} \priority{360} &  5.095 \priority{0}  \\ \hline
				\text{Average} &  \textbf{0.599} \priority{360} &  0.594 \priority{0} &  \textbf{2.314} \priority{360} &  2.923 \priority{0}  \\ \hline
			\end{tabular}
		\end{table*}
  
\subsection{Multi-Phase Training}
\label{Sec:MultiPhase_Training}
We train \ournerf{} in two phases.
The first phase accounts for 20\% of the training and uses a height map to jump-start the training process.
As noted in Deng et al. \cite{Depth_Supervised_NeRF}, we can use the height information to speed up the training and accuracy of a neural radiance field.
We acquire a height map via space carving \cite{SC} and use it to modify the computed density during the early stages of training.
\textcolor{Changes}{The height map used for depth supervision are created from the same images used to train the NeRF.
			The images used for evaluating the NeRF and the ground truth height map \textit{are not used} during the construction of the height map used for training.}
In the first phase of training, we replace $\rho$ with $\rho_m$, which is
\begin{equation}
	\label{Eq_Merged_rho}
	\rho_m\PAR{\vec{X}} = \Gamma \rho\PAR{\vec{X}} + \PAR{1-\Gamma}\rho_{H}\PAR{\vec{X}, \delta_{\vec{X}}, H},
\end{equation}
where $\Gamma$ linearly increases from 0 to 1 during the first phase, and $\rho_{H}$ is the density required for the neural radiance field to match the height map.

The value of $\rho_H$ depends on the height map $H$, the location within the model $\vec{X}$, and the distance between $\vec{X}$ and the next point along the ray, $\delta_{\vec{X}}$.
It is
\begin{subnumcases}{\rho_H\PAR{\vec{X}, \delta_{\vec{X}}, H} = }
	\frac{10}{\delta_{\vec{X}}} & $\vec{X} \text{ is at or below H}$ \\
	0 & $\vec{X} \text{ otherwise.}$ 
\end{subnumcases}
From Equation \ref{Eq:PE}, it is evident that $\rho_H$ encourages each point at or below a surface to have a $P_E \approx 1.0$.
In addition to $\rho_m$, we add a prior approximation loss
\begin{eqnarray}
	\label{Eq_Prior_Loss}
	L_p\PAR{\vec{X}} = L_{\alpha,c}\PAR{e^{-\rho\PAR{\vec{X}}\delta_{\vec{X}}} - e^{-\rho_h\PAR{\vec{X}, \delta_{\vec{X}}, H} \delta_{\vec{X}}}}.
\end{eqnarray} 
Equation \ref{Eq_Prior_Loss} ensures that the $\rho$ will be updated in the early stages of training despite having minimal impact on the rendered color when $\Gamma$ is near zero.

The second phase accounts for the remaining 80\% of training and ceases using $\rho_m$.
Furthermore, we no longer use the prior loss defined in Equation \ref{Eq_Prior_Loss} as the errors in our prior height map cause $\rho_H$ to be less accurate than $\rho$.
In addition, height maps cannot represent overhangs that may be present in the scene.

\subsection{Hyperparamter Tuning}

\textcolor{Changes}{The method we employ for hyperparameter tuning involves two parts.
		In the first part, we manually adjust hyperparameters to create decent results.
		In the second part, we employ Bayesian optimization as described in \cite{bayes_opt} to refine the hyperparameters.
		This process was run on OMA 248, which is not included in the results sections.
		We excluded OMA 248 as the hyperparameter tuning process involves feedback from the ground truth lidar, and thus all of the data associated with OMA 248 becomes training data.}
	
	\textcolor{Changes}{To evaluate the performance of a set of hyperparameters, we must combine the results of multiple metrics, measuring different qualities.
		To this end, we combine the SSIM, MAE, and seasonal stability into a single score, which is maximized during Bayesian optimization.
		This score is
		\begin{equation}
        \label{tune_eq}
			Score = \frac{SSIM}{SSIM_B} - \frac{MAE}{MAE_B} + \begin{cases}
				1 & EM < EM_B \\
				0 & else
			\end{cases}
		\end{equation}
		where $SSIM_B$, $MAE_B$, and $EM_B$ are the scores corresponding to the manually tuned parameters.
		In Equation \ref{tune_eq}, $EM$ refers to the worst earth-movers distance found after applying the seasonal stability process described in Section \ref{Sec:Season_Specific_and_Stable_Results}.
		Creating and evaluating \ournerf{} for a set of hyperparameters takes approximately eight hours.
		To speed up this process, we reduce the number of training steps from 50000 to 5000 and down-sample the images to keep the number of epochs consistent with the number of epochs when using 50000 steps.
		As a result, the training and evaluation for a set of hyperparameters is reduced to about 30 minutes.}

\section{Tests and Results}
\label{Sec:Results}

\label{sec:the_data}
We analyze the performance of \ournerf{} on areas of interest
from the \textit{2019 IEEE GRSS Data Fusion Contest}
\cite{Sat_data_2}, which contains images captured by Maxar WorldView-3
between 2014 and 2016.
\textcolor{Changes}{WorldView-3 captures 8-band visible and near-infrared images; however, \cite{Sat_data_2} provides pan-sharpened RGB images.
Images used in \ournerf{} undergo top-of-atmosphere correction to remove most of the variation caused by daily changes in atmospheric conditions.}
The images are of the city of Omaha, Nebraska,
USA and are RGB images of size 2048 by 2048 pixels covering an area of
approximately 580 by 580 meters.
However, we downsample the images to be 512 pixels by 512 pixels for training.
Table \ref{Tab:Data_Overview} contains a summary of each region.

\begin{table}[!htb]
	\centering
	\caption{\label{Tab:Data_Overview}Overview of data.
		For each area, four images are reserved for testing.}
	\begin{tabular}{|c|c|S[table-format=2.1, round-precision=1, round-mode=places]|}
		\hline
		\text{Area Index} & \text{Num. Imgs.} & \text{Height Range (m)} \\ \hline
		\text{OMA 042} & 39 & 38.339019775390625 \\ \hline
		\text{OMA 084} & 26 & 61.0782470703125 \\ \hline
		\text{OMA 132} & 41 & 31.942169189453125 \\ \hline
		\text{OMA 163} & 40 & 40.14947509765625 \\ \hline
		\text{OMA 203} & 43 & 51.653533935546875 \\ \hline
		\text{OMA 212} & 38 & 34.604248046875 \\ \hline
		\text{OMA 281} & 41 & 67.33929443359375 \\ \hline
		\text{OMA 374} & 30 & 59.4183349609375 \\ \hline
	\end{tabular}
\end{table}

\begin{figure*}[!t]
	\centering
	\includegraphics[width=1.0\linewidth]{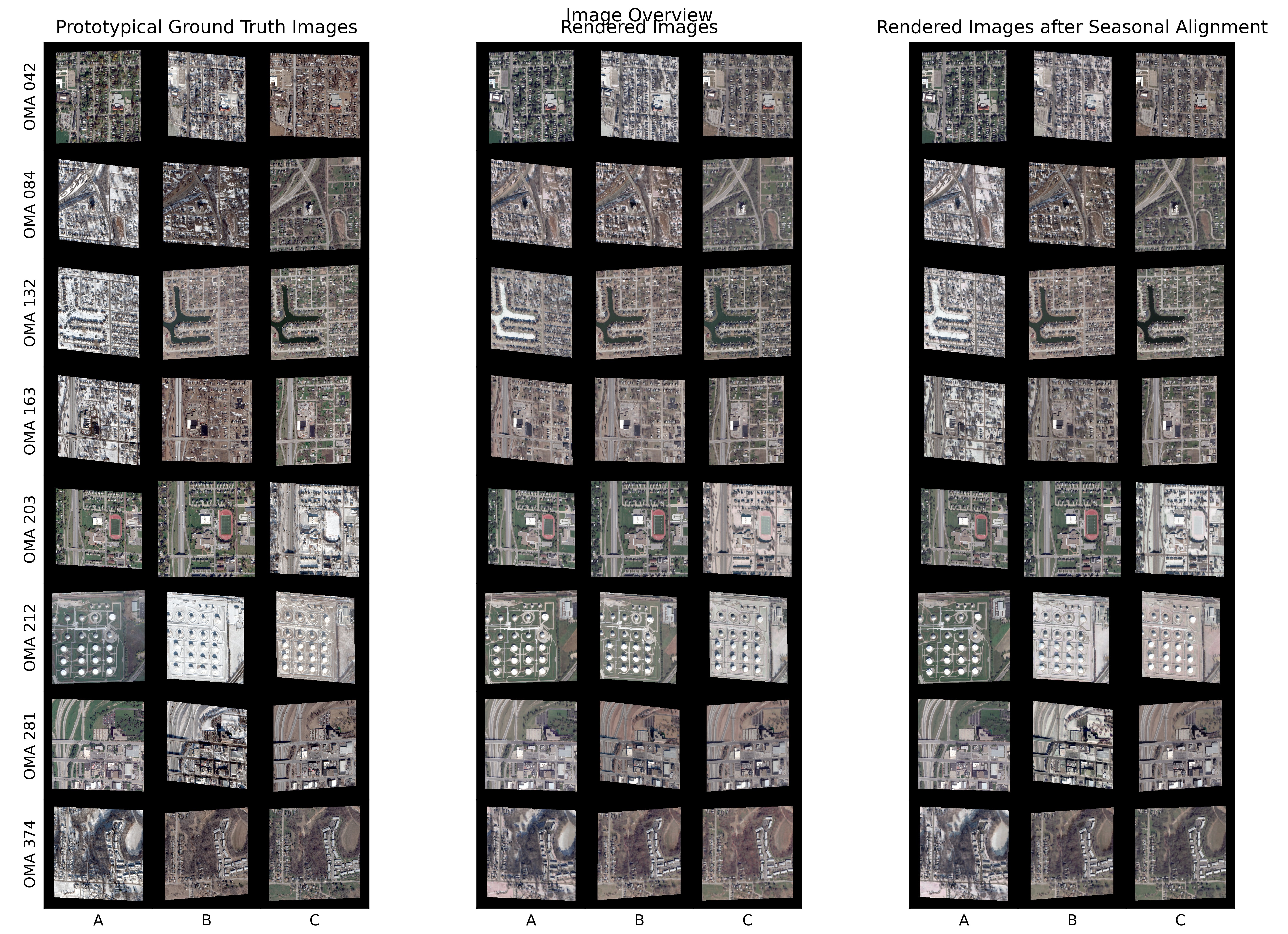}
	\caption{\label{Fig:Prototypical_Testing_Images}
		The left hyper-column contains ground truth images, the center hyper-column contains rendered images with direct time input, and the right hyper-column contains rendered images after seasonal alignment. }
\end{figure*}

\begin{figure*}[!t]
	\centering
	\includegraphics[width=1.0\linewidth]{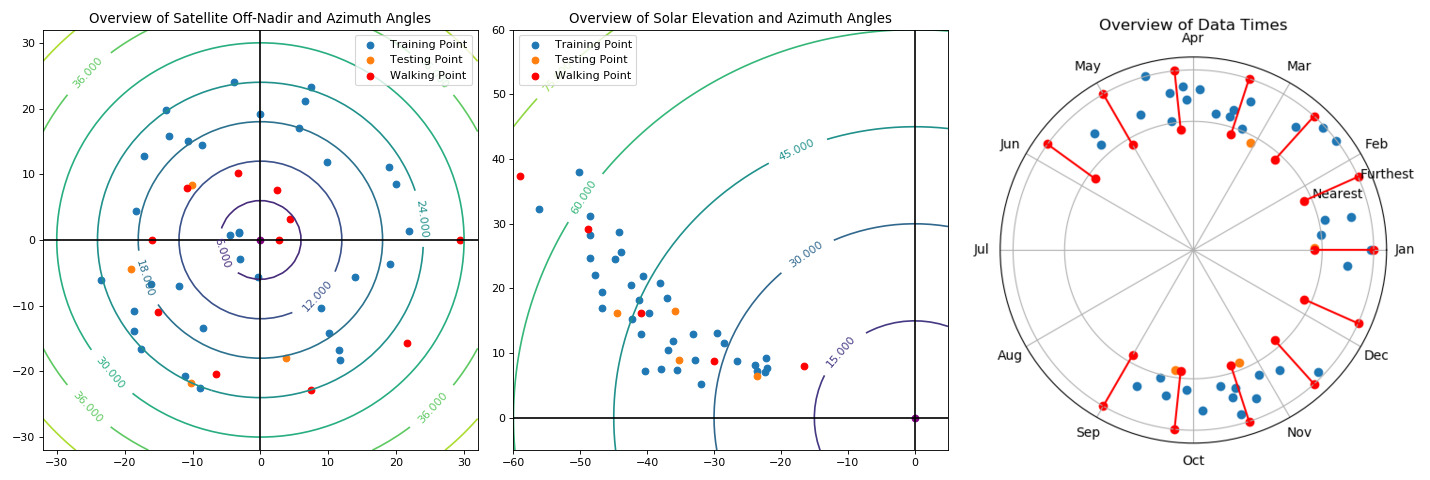}
	\caption{\label{Fig:Data_Example} Distribution of viewing angles, solar angles, and times for OMA 042.
		The satellite and solar angle graphs are on a polar scale, with distance from the origin measuring off-nadir and elevation angles.  The rotation from the positive x-axis measures the azimuth angle.
		The time plot uses rotation to reflect the time of the year.
        Distance from the origin indicates the sum of the solar and viewing angle difference between the training point and the closest testing point in terms of days apart.
        The smallest summed rotational difference is 23.4 degrees, and the largest is 187.0 degrees.
        The walking points occur at multiple times and viewing angles and are represented by a line, indicating that views near and far from the training points are used.}
\end{figure*}

Given a set of images, we withhold four for testing purposes and use
the remaining for training.  Three of these images represent
prototypical seasons, and the fourth is selected to ensure a wide range
of viewing and solar angles.  The prototypical seasonal images are
selected to ensure each testing set contains an image with large
amounts of snow, green foliage, and brown foliage.  The prototypical images are
shown in Fig. \ref{Fig:Prototypical_Testing_Images} with an example
distribution of the training data shown in
Fig. \ref{Fig:Data_Example}.  We use bundle-adjusted RPCs acquired by the method described in \cite{Raptor_Align} with initial height maps acquired by
the space carving process propounded in \cite{SC}.


\subsection{Novel View and Seasonal Variability Tests}
\label{Sec:Novel_View_Results}
To evaluate the quality of our model's novel view renderings, we use the Peak Signal-to-Noise Ratio (PSNR) and the Structural Similarity Index Measure (SSIM) \cite{SIMM} and compare generated images with images withheld from the training process.
In addition to applying these metrics to the output using the exact times provided in the satellite image's metadata, we also perform seasonal alignment.
These results are shown in Table \ref{Tab:Image_Quality}\footnote{For tests involving more than two cases, we include multicolored indicators.
			The amount of green shown in the indicator is proportional to the distance to the best result for the metric.}.

Seasons can change rapidly, and weather events sometimes occur unusually early or late in a season.
These sudden seasonal changes can cause the network to render an image containing seasonal characteristics that are valid for the time of the year but do not match the exact characteristics shown in the ground truth.
Seasonal alignment is accomplished by finding the time of year and the sky color such that the MSE between the rendered image and an image containing the desired seasonal features is minimized.
Examples of rendered images are shown in Fig. \ref{Fig:Prototypical_Testing_Images}.
OMA 084, OMA 212, and OMA 281 are times when seasonal alignment is very beneficial.

Examining the rendered images and image similarity scores provides evidence that \ournerf{} can construct images at novel view angles.
Selecting a wide range of seasons for the testing images allows us to check that seasonally independent features persist in the rendered images.
Objects, like buildings and roads, have their appearance modified by seasonal events throughout the year but are still visible and identifiable.
Additional evidence of this capability is provided in Sec.  \ref{Sec:Season_Specific_and_Stable_Results}.

\begin{figure*}[!t]
	\centering
	\includegraphics[width=1.0\linewidth]{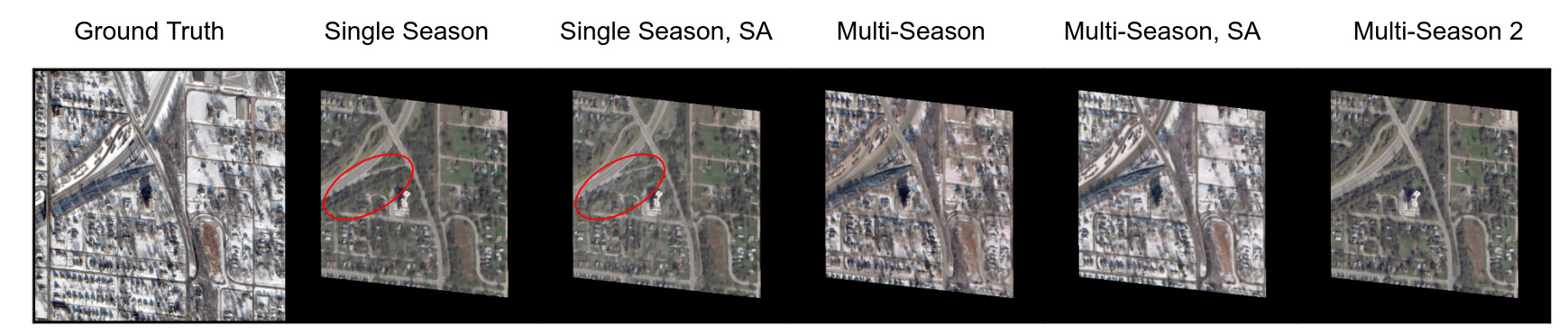}
	\caption{\label{Fig:Seasonal_Change_Imgs}
		This image is an example of how a single season limits the ability of NeRF to render novel views.
		Despite generating a high-quality rendering with accurate buildings and roads, the seasonal features in the single-season output are incorrect.
		Attempting seasonal alignment allows the sky color to change and the model to turn part of a forest into shadow (area circled in red) but does not allow the season to change due to the limitations of how shadows can affect the rendering.
		However, the multi-season approach generates an image with the correct buildings, roads, and seasonal features for a winter scene.
		Seasonal alignment improves these results by allowing the network to generate a winter scene with closer seasonal features to the ground truth.
		Also included on the far right is an image generated with seasonal features similar to the single-season model as evidence that the multi-season approach can generate multiple seasons with a single model.
	}
\end{figure*}

We consider five cases to examine the effectiveness of \ournerf{}'s components.
These cases are the entire implementation of \ournerf{} (Case A), \ournerf{} using S-NeRF's form of shadow prediction (Case B), \ournerf{} replacing Barron's loss with MSE loss (Case C), \ournerf{} without using a height map to guide the early stages of training (Case D), and \ournerf{} with only a single temporal class available (Case E).
An example of how the inclusion of multiple temporal classes alters the output is shown in Fig. \ref{Fig:Seasonal_Change_Imgs}.
From visual similarity scores shown in Table \ref{Tab:Image_Quality}, we conclude that allowing a model to account for multiple seasons drastically improves the quality of the rendered images.

\begin{table*}[!htb]
	\centering
	\caption{\label{Tab:Image_Quality}Results of comparing rendered images with testing images.  Included are directly rendered images and scores after seasonal alignment (SA).  Case A: Full Model.  Case B: Full Model, S-NeRF Solar loss.  Case C: MSE Loss.  Case D: No Height Map.  Case E: No Seasonal Adjustment.}
	\begin{tabular}{c c c c c c c c c c c }
			\hline
			\multicolumn{1}{|c|}{\text{Region}} & \multicolumn{5}{c|}{\text{PSNR $\uparrow$}} & \multicolumn{5}{c|}{\text{SSIM $\uparrow$}} \\ \hline
			\multicolumn{1}{|c|}{\text{Cases:}}& \multicolumn{1}{c}{\text{A}} & \multicolumn{1}{c}{\text{B}} & \multicolumn{1}{c}{\text{C}} & \multicolumn{1}{c}{\text{D}} & \multicolumn{1}{c|}{\text{E}} & \multicolumn{1}{c}{\text{A}} & \multicolumn{1}{c}{\text{B}} & \multicolumn{1}{c}{\text{C}} & \multicolumn{1}{c}{\text{D}} & \multicolumn{1}{c|}{\text{E}} \\ \hline
			\multicolumn{1}{|c|}{\text{OMA 042}} & 19.35 \priority{346}  &\textbf{19.43} \priority{360}  & 19.30 \priority{337}  & 19.22 \priority{323}  & \multicolumn{1}{c |}{17.45 \priority{0} } & 0.60 \priority{351}  &\textbf{0.60} \priority{360}  & 0.60 \priority{351}  & 0.59 \priority{330}  & \multicolumn{1}{c |}{0.48 \priority{0} }  \\ \hline
			\multicolumn{1}{|c|}{\text{OMA 084}} &\textbf{19.17} \priority{360}  & 18.46 \priority{249}  & 19.00 \priority{332}  & 19.12 \priority{352}  & \multicolumn{1}{c |}{16.85 \priority{0} } &\textbf{0.55} \priority{360}  & 0.52 \priority{300}  & 0.54 \priority{353}  & 0.53 \priority{324}  & \multicolumn{1}{c |}{0.35 \priority{0} }  \\ \hline
			\multicolumn{1}{|c|}{\text{OMA 132}} & 19.82 \priority{317}  & 19.84 \priority{320}  &\textbf{20.12} \priority{360}  & 20.12 \priority{359}  & \multicolumn{1}{c |}{17.59 \priority{0} } & 0.62 \priority{326}  & 0.61 \priority{293}  &\textbf{0.64} \priority{360}  & 0.63 \priority{348}  & \multicolumn{1}{c |}{0.51 \priority{0} }  \\ \hline
			\multicolumn{1}{|c|}{\text{OMA 163}} & 18.65 \priority{214}  & 18.78 \priority{240}  &\textbf{19.36} \priority{360}  & 18.61 \priority{205}  & \multicolumn{1}{c |}{17.60 \priority{0} } & 0.52 \priority{197}  & 0.54 \priority{277}  &\textbf{0.56} \priority{360}  & 0.52 \priority{216}  & \multicolumn{1}{c |}{0.46 \priority{0} }  \\ \hline
			\multicolumn{1}{|c|}{\text{OMA 203}} & 21.02 \priority{315}  & 21.21 \priority{342}  &\textbf{21.34} \priority{360}  & 20.39 \priority{225}  & \multicolumn{1}{c |}{18.80 \priority{0} } & 0.65 \priority{343}  & 0.65 \priority{355}  &\textbf{0.65} \priority{360}  & 0.63 \priority{296}  & \multicolumn{1}{c |}{0.55 \priority{0} }  \\ \hline
			\multicolumn{1}{|c|}{\text{OMA 212}} & 17.79 \priority{265}  & 18.12 \priority{311}  &\textbf{18.46} \priority{360}  & 17.65 \priority{244}  & \multicolumn{1}{c |}{15.93 \priority{0} } & 0.57 \priority{266}  & 0.59 \priority{352}  &\textbf{0.59} \priority{360}  & 0.56 \priority{210}  & \multicolumn{1}{c |}{0.51 \priority{0} }  \\ \hline
			\multicolumn{1}{|c|}{\text{OMA 281}} & 19.74 \priority{184}  & 19.96 \priority{245}  &\textbf{20.36} \priority{360}  & 19.91 \priority{233}  & \multicolumn{1}{c |}{19.08 \priority{0} } & 0.60 \priority{300}  & 0.61 \priority{339}  &\textbf{0.61} \priority{360}  & 0.60 \priority{278}  & \multicolumn{1}{c |}{0.55 \priority{0} }  \\ \hline
			\multicolumn{1}{|c|}{\text{OMA 374}} & 20.48 \priority{344}  & 20.33 \priority{310}  &\textbf{20.55} \priority{360}  & 20.20 \priority{283}  & \multicolumn{1}{c |}{18.91 \priority{0} } & 0.52 \priority{306}  & 0.53 \priority{337}  &\textbf{0.54} \priority{360}  & 0.52 \priority{298}  & \multicolumn{1}{c |}{0.45 \priority{0} }  \\ \hline
			\multicolumn{1}{|c|}{\text{Average}} & 19.50 \priority{305}  & 19.51 \priority{307}  &\textbf{19.81} \priority{360}  & 19.40 \priority{287}  & \multicolumn{1}{c |}{17.78 \priority{0} } & 0.58 \priority{317}  & 0.58 \priority{325}  &\textbf{0.59} \priority{360}  & 0.57 \priority{299}  & \multicolumn{1}{c |}{0.48 \priority{0} }  \\ \hline
			& & & & & & & & & &\\
			\cline{1-11}
			\multicolumn{1}{|c|}{\text{Region}} & \multicolumn{5}{c|}{\text{PSNR, SA $\uparrow$}} & \multicolumn{5}{c|}{\text{SSIM, SA $\uparrow$}} \\ \cline{1-11}
			\multicolumn{1}{|c|}{\text{Cases:}}& \multicolumn{1}{c}{\text{A}} & \multicolumn{1}{c}{\text{B}} & \multicolumn{1}{c}{\text{C}} & \multicolumn{1}{c}{\text{D}} & \multicolumn{1}{c|}{\text{E}} & \multicolumn{1}{c}{\text{A}} & \multicolumn{1}{c}{\text{B}} & \multicolumn{1}{c}{\text{C}} & \multicolumn{1}{c}{\text{D}} & \multicolumn{1}{c|}{\text{E}} \\ \cline{1-11}
			\multicolumn{1}{|c|}{\text{OMA 042}} & 19.73 \priority{341}  &\textbf{19.85} \priority{360}  & 19.83 \priority{355}  & 19.65 \priority{329}  & \multicolumn{1}{c |}{17.46 \priority{0} } & 0.60 \priority{355}  &\textbf{0.60} \priority{360}  & 0.60 \priority{356}  & 0.60 \priority{344}  & \multicolumn{1}{c |}{0.48 \priority{0} }  \\ \cline{1-11}
			\multicolumn{1}{|c|}{\text{OMA 084}} & 20.32 \priority{345}  & 20.25 \priority{337}  &\textbf{20.46} \priority{360}  & 20.04 \priority{316}  & \multicolumn{1}{c |}{16.96 \priority{0} } &\textbf{0.57} \priority{360}  & 0.56 \priority{336}  & 0.57 \priority{356}  & 0.54 \priority{310}  & \multicolumn{1}{c |}{0.35 \priority{0} }  \\ \cline{1-11}
			\multicolumn{1}{|c|}{\text{OMA 132}} & 21.21 \priority{337}  &\textbf{21.45} \priority{360}  & 21.27 \priority{343}  & 21.18 \priority{334}  & \multicolumn{1}{c |}{17.60 \priority{0} } &\textbf{0.67} \priority{360}  & 0.67 \priority{352}  & 0.67 \priority{353}  & 0.67 \priority{345}  & \multicolumn{1}{c |}{0.51 \priority{0} }  \\ \cline{1-11}
			\multicolumn{1}{|c|}{\text{OMA 163}} &\textbf{19.82} \priority{360}  & 19.76 \priority{351}  & 19.80 \priority{356}  & 19.59 \priority{322}  & \multicolumn{1}{c |}{17.61 \priority{0} } &\textbf{0.59} \priority{360}  & 0.57 \priority{321}  & 0.58 \priority{353}  & 0.56 \priority{296}  & \multicolumn{1}{c |}{0.46 \priority{0} }  \\ \cline{1-11}
			\multicolumn{1}{|c|}{\text{OMA 203}} & 21.44 \priority{330}  &\textbf{21.67} \priority{360}  & 21.63 \priority{355}  & 21.23 \priority{304}  & \multicolumn{1}{c |}{18.83 \priority{0} } & 0.66 \priority{350}  &\textbf{0.66} \priority{360}  & 0.65 \priority{342}  & 0.64 \priority{314}  & \multicolumn{1}{c |}{0.55 \priority{0} }  \\ \cline{1-11}
			\multicolumn{1}{|c|}{\text{OMA 212}} & 20.26 \priority{299}  &\textbf{21.09} \priority{360}  & 20.47 \priority{314}  & 20.32 \priority{303}  & \multicolumn{1}{c |}{16.20 \priority{0} } & 0.61 \priority{277}  &\textbf{0.64} \priority{360}  & 0.63 \priority{326}  & 0.61 \priority{284}  & \multicolumn{1}{c |}{0.51 \priority{0} }  \\ \cline{1-11}
			\multicolumn{1}{|c|}{\text{OMA 281}} & 20.99 \priority{323}  &\textbf{21.19} \priority{360}  & 21.16 \priority{353}  & 21.03 \priority{332}  & \multicolumn{1}{c |}{19.15 \priority{0} } & 0.64 \priority{357}  &\textbf{0.64} \priority{360}  & 0.63 \priority{325}  & 0.62 \priority{297}  & \multicolumn{1}{c |}{0.55 \priority{0} }  \\ \cline{1-11}
			\multicolumn{1}{|c|}{\text{OMA 374}} & 21.28 \priority{337}  &\textbf{21.44} \priority{360}  & 21.43 \priority{358}  & 20.91 \priority{284}  & \multicolumn{1}{c |}{18.95 \priority{0} } & 0.53 \priority{320}  & 0.54 \priority{334}  &\textbf{0.54} \priority{360}  & 0.53 \priority{295}  & \multicolumn{1}{c |}{0.45 \priority{0} }  \\ \cline{1-11}
			\multicolumn{1}{|c|}{\text{Average}} & 20.63 \priority{335}  &\textbf{20.84} \priority{360}  & 20.75 \priority{350}  & 20.49 \priority{318}  & \multicolumn{1}{c |}{17.85 \priority{0} } & 0.61 \priority{356}  & 0.61 \priority{358}  &\textbf{0.61} \priority{360}  & 0.60 \priority{323}  & \multicolumn{1}{c |}{0.48 \priority{0} }  \\ \cline{1-11}
		\end{tabular}
\end{table*}

\subsection{Height Map and Shadow Mask Tests}
\label{Sec:HM_Results}
\begin{figure}[!t]
	\centering
	\includegraphics[width=1.0\linewidth]{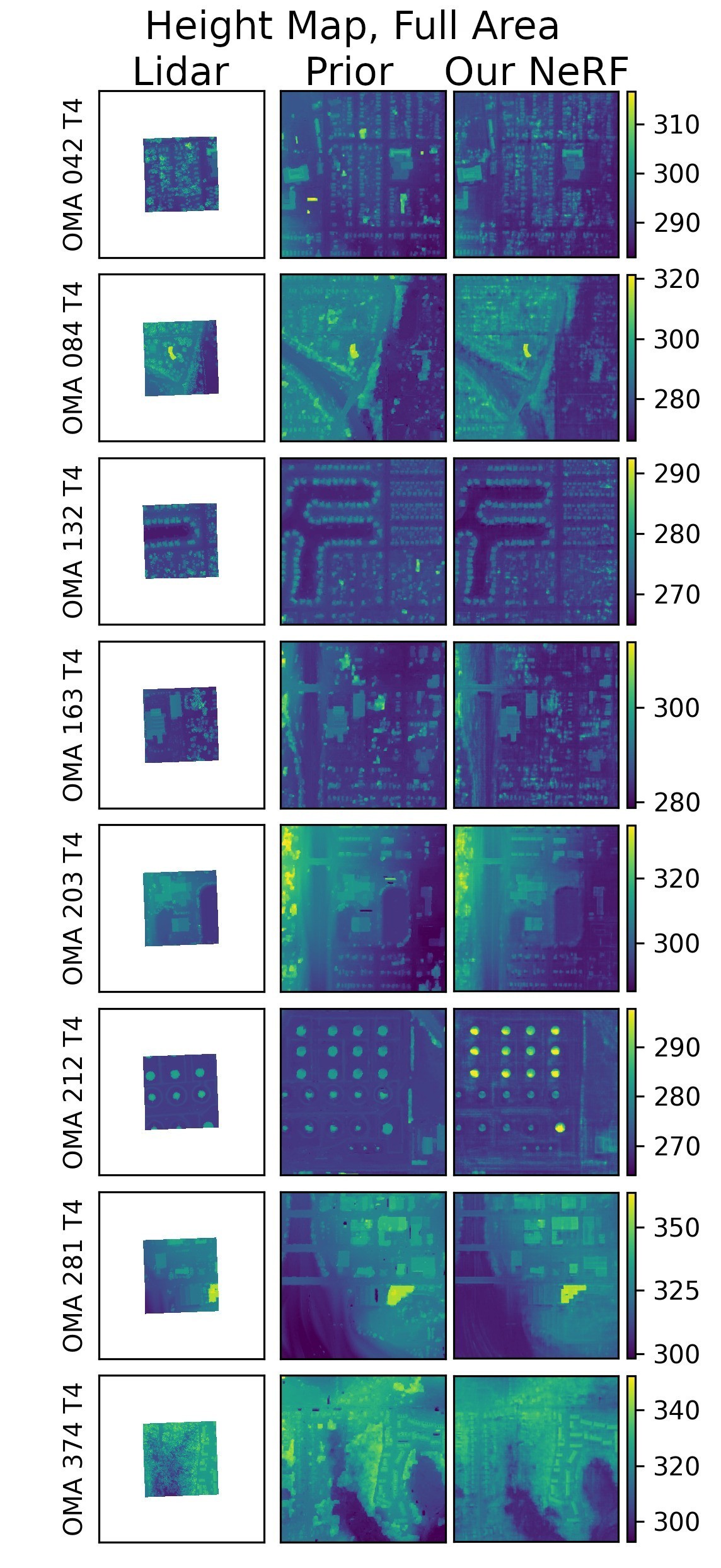}
	\caption{\label{Fig:HM_Img} Height maps of lidar data, where available (left), prior height map (center), and \ournerf{}'s height map (right).}
\end{figure}

To evaluate the height map, we compute the mean absolute height error (MAE), as described in \cite{Metrics_3D_and_data}, relative to lidar.
Examples of the height maps rendered by \ournerf{} process are shown in Fig. \ref{Fig:HM_Img}.
Table \ref{Tab:HM_Quality} contains a quantitative comparison of our height maps with lidar data.
We considered the same cases as in Section \ref{Sec:Novel_View_Results} for the evaluation of the height map.
As with image quality, the inclusion of seasonal variation resulted in a substantial improvement to the height map.
In addition, the inclusion of a prior is a significant factor in the quality of the final height map generated by \ournerf{}.

\begin{figure}[!t]
	\centering
	\includegraphics[width=0.75\linewidth]{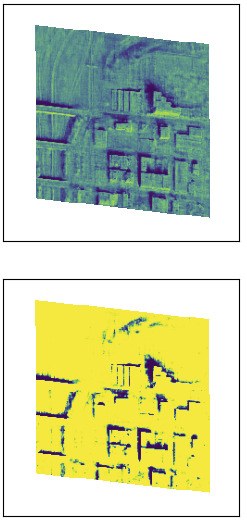}
	\caption{\label{Fig:Analog_Shadows} Example of application of Equation \ref{Eq:Shadow_Mask} (bottom), as well as the image before sigmoid, scaling via $\kappa$, and shifting via $\mu$ are used (top).
			Both images are in the range of $\left\lbrack 0,1\right\rbrack$.}
\end{figure}

\begin{figure}[!t]
	\centering
	\includegraphics[width=1.0\linewidth]{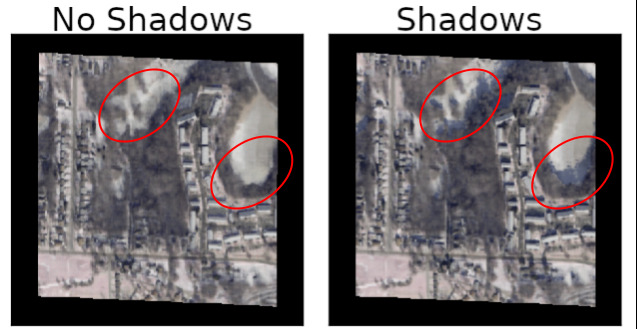}
	\caption{\label{Fig:Shadow_Effects} Example of shadow rendering on OMA 374.
 Results of rendered images without shadow mask (left) and with shadow mask (right).
 Regions circled in red contain shadows that are not rendered until after applying the shadow mask.}
\end{figure}

\begin{figure}[tp]
	\centering
	\includegraphics[height=.95\textheight]{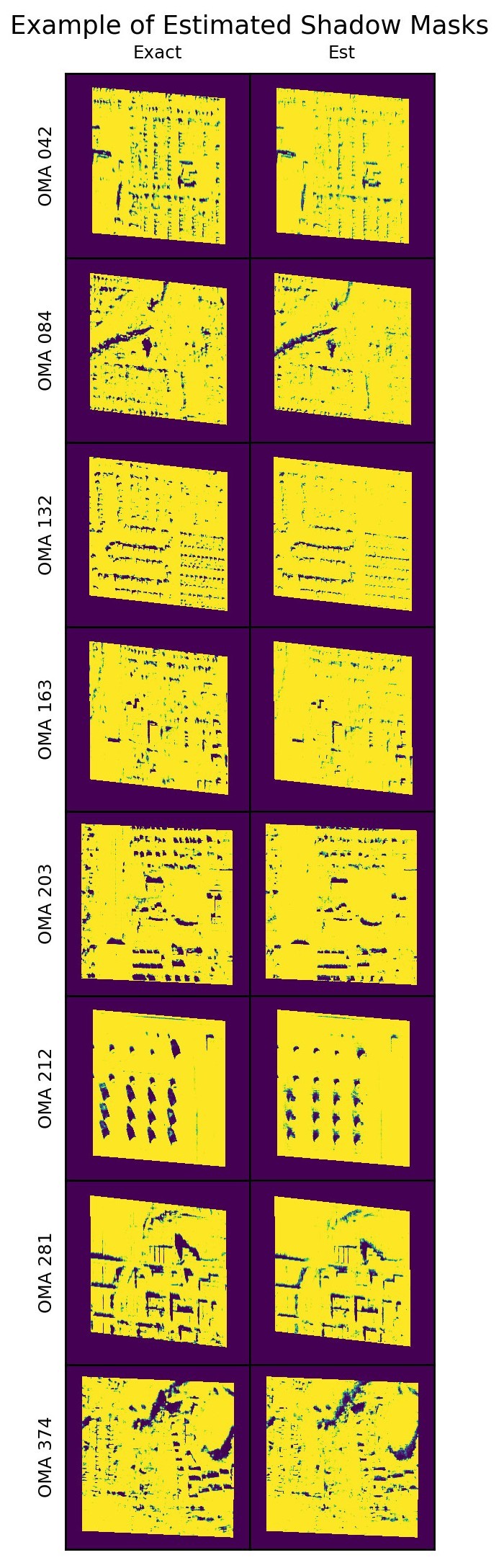}
	\caption{\label{Fig:Shadow_Img} Examples of estimated shadow masks (right) compared to exact shadow masks (left)}
\end{figure}

To measure the quality of our predicted shadow masks, we compute an exact shadow mask using Equation \ref{Eq:Shadow_Mask} by replacing $V\PAR{\vec{X}_i, \vec{w}}$ with the exact solar visibility computed along the ray by Equation \ref{Eq:PV}.
In theory, we could always use the exact value to compute the shadow mask instead of using a network to approximate it.
However, this is not practical as using the exact computation to render a ray with $n$ points is $O\PAR{n^2}$.
Instead, if the approximate computation is used, rendering is a $O\PAR{n}$ operation.
Examples of the predicted shadow mask and the exact shadow mask are shown in Fig. \ref{Fig:Shadow_Img} with numerical results shown in Table \ref{Tab:Shadow_Quality}.
\textcolor{Changes}{In addition Fig. \ref{Fig:Analog_Shadows} shows an example of the shadow mask from Equation \ref{Eq:Shadow_Mask} before applying the sigmoid function.}
	\textcolor{Changes}{In Fig. \ref{Fig:Shadow_Effects}, we show an image before and after the shadow mask has been applied.}
\textcolor{Changes}{We compare the estimated shadow mask to the exact shadow mask instead of the shadow mask from the lidar DSM as shadows are predicted entirely based on the model density.
Errors in the model density will result in errors in the shadows and create ambiguity regarding the performance of the estimated shadow mask, as it is unclear if a loss in accuracy is due to poor approximation of the exact shadows or problems in the density model.}

\begin{figure}[!t]
	\centering
	\includegraphics[width=1.0\linewidth]{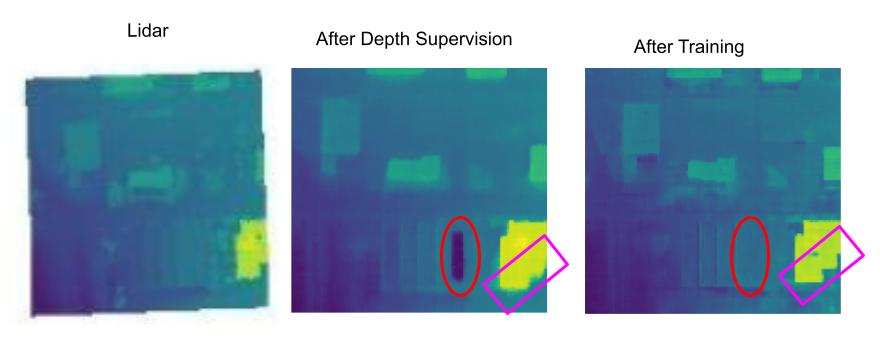}
	\caption{\label{Fig:HM_with_DSM}  Lidar data of OMA 281 (left). Intermediate height map generated by \ournerf{} at the end of depth supervision (center).
				Final height map generated by \ournerf{} (right).
				Note the gauge (circled in red) from the depth supervision and the fuzzy building border (in pink rectangle).
				These errors in the height map are from the prior, and the training process corrects them after depth supervision stops.}
\end{figure}

\textcolor{Changes}{As with \cite{Depth_Supervised_NeRF} and \cite{Sat_NeRF}, the inclusion of prior height information drastically improves the quality of the model generated by \ournerf{}.
			However, we also compare the quality of the model learned with the prior height map with the quality of the prior height map itself.
			We note that in most cases, \ournerf{} results in a height map superior to the one provided during training.
			As shown in Fig. \ref{Fig:HM_with_DSM}, \ournerf{} initially learns height information from the prior height map, including some of the errors in the prior height map.
			However, in the later phase of the training, when the prior height map is no longer being used, the training process can correct the errors in the prior.
			Correcting these errors results in a superior height map from the NeRF compared to the prior provided to the NeRF for training.}

\begin{table*}[!htb]
	\centering
	\caption{\label{Tab:HM_Quality}Quality of height maps relative to lidar data.  Highlighted in green is the best score per metric for each region.  Case Prior: Space Carved Height Map.  Case A: Full Model.  Case B: Full Model, S-NeRF Solar loss.  Case C: MSE Loss.  Case D: No Height Map.  Case E: No Seasonal Adjustment.}
	\begin{tabular}{c c c c c c c c c c c c c }
			\hline
			\multicolumn{1}{|c|}{\text{Reg}} & \multicolumn{6}{c|}{\text{MAE $\downarrow$}} & \multicolumn{6}{c|}{\text{RMSE $\downarrow$}} \\ \hline
			\multicolumn{1}{|c|}{}& \multicolumn{1}{c}{\text{Prior}} & \multicolumn{1}{c}{\text{A}} & \multicolumn{1}{c}{\text{B}} & \multicolumn{1}{c}{\text{C}} & \multicolumn{1}{c}{\text{D}} & \multicolumn{1}{c|}{\text{E}} & \multicolumn{1}{c}{\text{Prior}} & \multicolumn{1}{c}{\text{A}} & \multicolumn{1}{c}{\text{B}} & \multicolumn{1}{c}{\text{C}} & \multicolumn{1}{c}{\text{D}} & \multicolumn{1}{c|}{\text{E}} \\ \hline
			\multicolumn{1}{|c|}{\text{042}} & 2.92 \priority{0}  & 2.47 \priority{291}  &\textbf{2.37} \priority{360}  & 2.41 \priority{331}  & 2.43 \priority{320}  & \multicolumn{1}{c |}{2.79 \priority{83} } & 4.08 \priority{0}  & 3.56 \priority{283}  & 3.44 \priority{353}  & 3.47 \priority{334}  &\textbf{3.42} \priority{360}  & \multicolumn{1}{c |}{3.79 \priority{155} }  \\ \hline
			\multicolumn{1}{|c|}{\text{084}} & 3.34 \priority{305}  &\textbf{2.82} \priority{360}  & 3.11 \priority{329}  & 2.90 \priority{352}  & 3.94 \priority{241}  & \multicolumn{1}{c |}{6.22 \priority{0} } & 5.06 \priority{272}  &\textbf{4.13} \priority{360}  & 4.60 \priority{315}  & 4.39 \priority{335}  & 5.14 \priority{264}  & \multicolumn{1}{c |}{7.95 \priority{0} }  \\ \hline
			\multicolumn{1}{|c|}{\text{132}} &\textbf{1.20} \priority{360}  & 1.22 \priority{347}  & 1.26 \priority{315}  & 1.23 \priority{342}  & 1.46 \priority{164}  & \multicolumn{1}{c |}{1.67 \priority{0} } & 2.13 \priority{264}  &\textbf{2.02} \priority{360}  & 2.10 \priority{292}  & 2.07 \priority{318}  & 2.06 \priority{322}  & \multicolumn{1}{c |}{2.44 \priority{0} }  \\ \hline
			\multicolumn{1}{|c|}{\text{163}} & 1.80 \priority{227}  &\textbf{1.43} \priority{360}  & 1.68 \priority{268}  & 1.88 \priority{197}  & 2.04 \priority{138}  & \multicolumn{1}{c |}{2.42 \priority{0} } & 2.75 \priority{159}  &\textbf{2.31} \priority{360}  & 2.69 \priority{187}  & 2.61 \priority{222}  & 2.77 \priority{151}  & \multicolumn{1}{c |}{3.10 \priority{0} }  \\ \hline
			\multicolumn{1}{|c|}{\text{203}} & 1.37 \priority{245}  & 1.17 \priority{321}  & 1.44 \priority{218}  &\textbf{1.07} \priority{360}  & 2.02 \priority{0}  & \multicolumn{1}{c |}{1.87 \priority{55} } & 2.61 \priority{74}  & 1.92 \priority{320}  & 2.45 \priority{132}  &\textbf{1.81} \priority{360}  & 2.82 \priority{0}  & \multicolumn{1}{c |}{2.82 \priority{1} }  \\ \hline
			\multicolumn{1}{|c|}{\text{212}} &\textbf{0.64} \priority{360}  & 0.94 \priority{141}  & 0.84 \priority{213}  & 0.91 \priority{159}  & 1.11 \priority{17}  & \multicolumn{1}{c |}{1.13 \priority{0} } &\textbf{1.44} \priority{360}  & 1.76 \priority{149}  & 1.59 \priority{261}  & 1.69 \priority{198}  & 1.99 \priority{0}  & \multicolumn{1}{c |}{1.96 \priority{19} }  \\ \hline
			\multicolumn{1}{|c|}{\text{281}} & 1.77 \priority{275}  &\textbf{1.46} \priority{360}  & 1.51 \priority{347}  & 1.61 \priority{319}  & 2.79 \priority{0}  & \multicolumn{1}{c |}{2.65 \priority{36} } & 3.32 \priority{137}  &\textbf{2.46} \priority{360}  & 2.52 \priority{344}  & 2.67 \priority{304}  & 3.86 \priority{0}  & \multicolumn{1}{c |}{3.66 \priority{51} }  \\ \hline
			\multicolumn{1}{|c|}{\text{374}} & 5.27 \priority{141}  & 4.44 \priority{327}  &\textbf{4.29} \priority{360}  & 4.52 \priority{308}  & 5.22 \priority{153}  & \multicolumn{1}{c |}{5.91 \priority{0} } & 6.75 \priority{131}  & 5.88 \priority{333}  &\textbf{5.77} \priority{360}  & 5.90 \priority{328}  & 6.49 \priority{192}  & \multicolumn{1}{c |}{7.32 \priority{0} }  \\ \hline
			\multicolumn{1}{|c|}{\text{Avg}} & 2.29 \priority{262}  &\textbf{1.99} \priority{360}  & 2.06 \priority{337}  & 2.07 \priority{336}  & 2.62 \priority{151}  & \multicolumn{1}{c |}{3.08 \priority{0} } & 3.52 \priority{196}  &\textbf{3.01} \priority{360}  & 3.14 \priority{315}  & 3.08 \priority{337}  & 3.57 \priority{179}  & \multicolumn{1}{c |}{4.13 \priority{0} }  \\ \hline
			& & & & & & & & & & & &\\
			\cline{1-13}
			\multicolumn{1}{|c|}{\text{Reg}} & \multicolumn{6}{c|}{\text{Percent within 1 m $\uparrow$}} & \multicolumn{6}{c|}{\text{Median Error $\downarrow$}} \\ \cline{1-13}
			\multicolumn{1}{|c|}{}& \multicolumn{1}{c}{\text{Prior}} & \multicolumn{1}{c}{\text{A}} & \multicolumn{1}{c}{\text{B}} & \multicolumn{1}{c}{\text{C}} & \multicolumn{1}{c}{\text{D}} & \multicolumn{1}{c|}{\text{E}} & \multicolumn{1}{c}{\text{Prior}} & \multicolumn{1}{c}{\text{A}} & \multicolumn{1}{c}{\text{B}} & \multicolumn{1}{c}{\text{C}} & \multicolumn{1}{c}{\text{D}} & \multicolumn{1}{c|}{\text{E}} \\ \cline{1-13}
			\multicolumn{1}{|c|}{\text{042}} & 0.17 \priority{0}  & 0.26 \priority{223}  & 0.30 \priority{325}  &\textbf{0.31} \priority{360}  & 0.31 \priority{343}  & \multicolumn{1}{c |}{0.25 \priority{211} } & 2.22 \priority{0}  & 1.61 \priority{328}  &\textbf{1.55} \priority{360}  & 1.63 \priority{318}  & 1.71 \priority{272}  & \multicolumn{1}{c |}{2.16 \priority{35} }  \\ \cline{1-13}
			\multicolumn{1}{|c|}{\text{084}} &\textbf{0.34} \priority{360}  & 0.30 \priority{291}  & 0.28 \priority{257}  & 0.32 \priority{332}  & 0.16 \priority{48}  & \multicolumn{1}{c |}{0.13 \priority{0} } & 2.27 \priority{293}  & 1.82 \priority{343}  & 2.06 \priority{316}  &\textbf{1.68} \priority{360}  & 3.14 \priority{196}  & \multicolumn{1}{c |}{4.90 \priority{0} }  \\ \cline{1-13}
			\multicolumn{1}{|c|}{\text{132}} &\textbf{0.67} \priority{360}  & 0.61 \priority{265}  & 0.60 \priority{248}  & 0.63 \priority{289}  & 0.47 \priority{8}  & \multicolumn{1}{c |}{0.47 \priority{0} } &\textbf{0.63} \priority{360}  & 0.73 \priority{280}  & 0.75 \priority{261}  & 0.69 \priority{311}  & 1.08 \priority{12}  & \multicolumn{1}{c |}{1.10 \priority{0} }  \\ \cline{1-13}
			\multicolumn{1}{|c|}{\text{163}} & 0.41 \priority{167}  &\textbf{0.57} \priority{360}  & 0.48 \priority{258}  & 0.37 \priority{110}  & 0.35 \priority{94}  & \multicolumn{1}{c |}{0.28 \priority{0} } & 1.17 \priority{265}  &\textbf{0.84} \priority{360}  & 1.04 \priority{302}  & 1.41 \priority{194}  & 1.55 \priority{153}  & \multicolumn{1}{c |}{2.07 \priority{0} }  \\ \cline{1-13}
			\multicolumn{1}{|c|}{\text{203}} & 0.56 \priority{216}  & 0.65 \priority{321}  & 0.56 \priority{214}  &\textbf{0.68} \priority{360}  & 0.38 \priority{0}  & \multicolumn{1}{c |}{0.46 \priority{100} } & 0.86 \priority{250}  & 0.64 \priority{343}  & 0.89 \priority{235}  &\textbf{0.61} \priority{360}  & 1.44 \priority{0}  & \multicolumn{1}{c |}{1.12 \priority{135} }  \\ \cline{1-13}
			\multicolumn{1}{|c|}{\text{212}} &\textbf{0.85} \priority{360}  & 0.73 \priority{144}  & 0.76 \priority{200}  & 0.76 \priority{198}  & 0.67 \priority{28}  & \multicolumn{1}{c |}{0.65 \priority{0} } &\textbf{0.25} \priority{360}  & 0.45 \priority{200}  & 0.41 \priority{236}  & 0.49 \priority{166}  & 0.68 \priority{15}  & \multicolumn{1}{c |}{0.70 \priority{0} }  \\ \cline{1-13}
			\multicolumn{1}{|c|}{\text{281}} & 0.54 \priority{357}  &\textbf{0.54} \priority{360}  & 0.54 \priority{352}  & 0.50 \priority{312}  & 0.27 \priority{11}  & \multicolumn{1}{c |}{0.26 \priority{0} } & 0.90 \priority{354}  &\textbf{0.89} \priority{360}  & 0.91 \priority{351}  & 0.99 \priority{327}  & 2.02 \priority{7}  & \multicolumn{1}{c |}{2.04 \priority{0} }  \\ \cline{1-13}
			\multicolumn{1}{|c|}{\text{374}} & 0.12 \priority{73}  & 0.18 \priority{252}  &\textbf{0.21} \priority{360}  & 0.14 \priority{154}  & 0.13 \priority{106}  & \multicolumn{1}{c |}{0.09 \priority{0} } & 4.21 \priority{156}  & 3.23 \priority{336}  &\textbf{3.11} \priority{360}  & 3.31 \priority{322}  & 4.39 \priority{123}  & \multicolumn{1}{c |}{5.06 \priority{0} }  \\ \cline{1-13}
			\multicolumn{1}{|c|}{\text{Avg}} & 0.46 \priority{307}  &\textbf{0.48} \priority{360}  & 0.47 \priority{330}  & 0.46 \priority{325}  & 0.34 \priority{39}  & \multicolumn{1}{c |}{0.32 \priority{0} } & 1.56 \priority{267}  &\textbf{1.28} \priority{360}  & 1.34 \priority{339}  & 1.35 \priority{336}  & 2.00 \priority{126}  & \multicolumn{1}{c |}{2.39 \priority{0} }  \\ \cline{1-13}
		\end{tabular}
\end{table*}

%Note that \ournerf{} needs better shadow recall, resulting in our approach incorrectly classifying areas as not in shadow.
%As a result, the dark color in these regions is not caused by the sun but rather assumed to be the albedo color of the region.
%The model's inability to recall the shadows can cause ``false shadows'' to appear in the image.
%However, changing from MSE loss to Barron's loss results in an average improvement of 10\% for shadow recall.

\begin{table*}[!htb]
	\centering
	\caption{Quantitative overview of EM distance when time is constant. Case A: Full Model.  Case B: Full Model, S-NeRF Solar loss.  Case C: MSE Loss.  Case D: No Height Map.  Case E: No Seasonal Adjustment.
		While Case D contains the best shadow prediction, it also performs poorly for height map quality.}
	\begin{tabular}{c c c c c c c c c c c }
			\hline
			\multicolumn{1}{|c|}{\text{Region}} & \multicolumn{5}{c|}{\text{Accuracy $\uparrow$}} & \multicolumn{5}{c|}{\text{Sun F1 $\uparrow$}} \\ \hline
			\multicolumn{1}{|c|}{\text{Cases:}}& \multicolumn{1}{c}{\text{A}} & \multicolumn{1}{c}{\text{B}} & \multicolumn{1}{c}{\text{C}} & \multicolumn{1}{c}{\text{D}} & \multicolumn{1}{c|}{\text{E}} & \multicolumn{1}{c}{\text{A}} & \multicolumn{1}{c}{\text{B}} & \multicolumn{1}{c}{\text{C}} & \multicolumn{1}{c}{\text{D}} & \multicolumn{1}{c|}{\text{E}} \\ \hline
			\multicolumn{1}{|c|}{\text{OMA 042}} & 0.94 \priority{172}  & 0.94 \priority{197}  & 0.93 \priority{152}  &\textbf{0.97} \priority{360}  & \multicolumn{1}{c |}{0.91 \priority{0} } & 0.97 \priority{174}  & 0.97 \priority{203}  & 0.96 \priority{161}  &\textbf{0.98} \priority{360}  & \multicolumn{1}{c |}{0.95 \priority{0} }  \\ \hline
			\multicolumn{1}{|c|}{\text{OMA 084}} & 0.94 \priority{37}  & 0.95 \priority{70}  & 0.94 \priority{34}  &\textbf{0.98} \priority{360}  & \multicolumn{1}{c |}{0.94 \priority{0} } & 0.96 \priority{200}  & 0.98 \priority{265}  & 0.95 \priority{151}  &\textbf{1.00} \priority{360}  & \multicolumn{1}{c |}{0.91 \priority{0} }  \\ \hline
			\multicolumn{1}{|c|}{\text{OMA 132}} & 0.94 \priority{184}  & 0.94 \priority{168}  & 0.94 \priority{169}  &\textbf{0.98} \priority{360}  & \multicolumn{1}{c |}{0.91 \priority{0} } & 0.97 \priority{136}  & 0.97 \priority{139}  & 0.96 \priority{0}  &\textbf{0.99} \priority{360}  & \multicolumn{1}{c |}{0.97 \priority{35} }  \\ \hline
			\multicolumn{1}{|c|}{\text{OMA 163}} & 0.94 \priority{115}  & 0.95 \priority{175}  & 0.95 \priority{159}  &\textbf{0.98} \priority{360}  & \multicolumn{1}{c |}{0.92 \priority{0} } & 0.97 \priority{39}  & 0.98 \priority{188}  & 0.98 \priority{133}  &\textbf{0.99} \priority{360}  & \multicolumn{1}{c |}{0.97 \priority{0} }  \\ \hline
			\multicolumn{1}{|c|}{\text{OMA 203}} & 0.95 \priority{178}  & 0.93 \priority{103}  & 0.96 \priority{236}  &\textbf{0.98} \priority{360}  & \multicolumn{1}{c |}{0.91 \priority{0} } & 0.96 \priority{123}  & 0.96 \priority{112}  & 0.96 \priority{58}  &\textbf{0.99} \priority{360}  & \multicolumn{1}{c |}{0.95 \priority{0} }  \\ \hline
			\multicolumn{1}{|c|}{\text{OMA 212}} & 0.97 \priority{228}  & 0.98 \priority{295}  & 0.97 \priority{227}  &\textbf{0.99} \priority{360}  & \multicolumn{1}{c |}{0.93 \priority{0} } & 0.98 \priority{225}  & 0.99 \priority{318}  & 0.99 \priority{270}  &\textbf{0.99} \priority{360}  & \multicolumn{1}{c |}{0.97 \priority{0} }  \\ \hline
			\multicolumn{1}{|c|}{\text{OMA 281}} & 0.97 \priority{205}  & 0.96 \priority{136}  & 0.98 \priority{250}  &\textbf{1.00} \priority{360}  & \multicolumn{1}{c |}{0.93 \priority{0} } & 0.97 \priority{192}  & 0.96 \priority{151}  & 0.97 \priority{202}  &\textbf{0.99} \priority{360}  & \multicolumn{1}{c |}{0.94 \priority{0} }  \\ \hline
			\multicolumn{1}{|c|}{\text{OMA 374}} & 0.95 \priority{173}  & 0.92 \priority{1}  & 0.95 \priority{183}  &\textbf{0.98} \priority{360}  & \multicolumn{1}{c |}{0.92 \priority{0} } & 0.96 \priority{128}  & 0.95 \priority{71}  & 0.97 \priority{168}  &\textbf{1.00} \priority{360}  & \multicolumn{1}{c |}{0.94 \priority{0} }  \\ \hline
			\multicolumn{1}{|c|}{\text{Average}} & 0.95 \priority{167}  & 0.95 \priority{144}  & 0.95 \priority{183}  &\textbf{0.98} \priority{360}  & \multicolumn{1}{c |}{0.92 \priority{0} } & 0.97 \priority{158}  & 0.97 \priority{181}  & 0.97 \priority{143}  &\textbf{0.99} \priority{360}  & \multicolumn{1}{c |}{0.95 \priority{0} }  \\ \hline
			& & & & & & & & & &\\
			\cline{1-11}
			\multicolumn{1}{|c|}{\text{Region}} & \multicolumn{5}{c|}{\text{Shadow Precision $\uparrow$}} & \multicolumn{5}{c|}{\text{Shadow Recall $\uparrow$}} \\ \cline{1-11}
			\multicolumn{1}{|c|}{\text{Cases:}}& \multicolumn{1}{c}{\text{A}} & \multicolumn{1}{c}{\text{B}} & \multicolumn{1}{c}{\text{C}} & \multicolumn{1}{c}{\text{D}} & \multicolumn{1}{c|}{\text{E}} & \multicolumn{1}{c}{\text{A}} & \multicolumn{1}{c}{\text{B}} & \multicolumn{1}{c}{\text{C}} & \multicolumn{1}{c}{\text{D}} & \multicolumn{1}{c|}{\text{E}} \\ \cline{1-11}
			\multicolumn{1}{|c|}{\text{OMA 042}} & 0.77 \priority{237}  &\textbf{0.84} \priority{360}  & 0.78 \priority{260}  & 0.80 \priority{299}  & \multicolumn{1}{c |}{0.65 \priority{0} } & 0.47 \priority{309}  & 0.37 \priority{95}  & 0.32 \priority{0}  &\textbf{0.49} \priority{360}  & \multicolumn{1}{c |}{0.42 \priority{200} }  \\ \cline{1-11}
			\multicolumn{1}{|c|}{\text{OMA 084}} & 0.77 \priority{193}  & 0.70 \priority{119}  & 0.74 \priority{159}  &\textbf{0.91} \priority{360}  & \multicolumn{1}{c |}{0.61 \priority{0} } & 0.53 \priority{98}  & 0.41 \priority{0}  & 0.49 \priority{66}  &\textbf{0.85} \priority{360}  & \multicolumn{1}{c |}{0.71 \priority{240} }  \\ \cline{1-11}
			\multicolumn{1}{|c|}{\text{OMA 132}} & 0.82 \priority{238}  & 0.82 \priority{236}  & 0.76 \priority{116}  &\textbf{0.87} \priority{360}  & \multicolumn{1}{c |}{0.71 \priority{0} } & 0.48 \priority{255}  & 0.33 \priority{63}  & 0.41 \priority{172}  &\textbf{0.56} \priority{360}  & \multicolumn{1}{c |}{0.28 \priority{0} }  \\ \cline{1-11}
			\multicolumn{1}{|c|}{\text{OMA 163}} & 0.79 \priority{236}  & 0.88 \priority{351}  & 0.81 \priority{261}  &\textbf{0.89} \priority{360}  & \multicolumn{1}{c |}{0.62 \priority{0} } & 0.47 \priority{202}  & 0.31 \priority{0}  & 0.31 \priority{0}  &\textbf{0.58} \priority{360}  & \multicolumn{1}{c |}{0.41 \priority{130} }  \\ \cline{1-11}
			\multicolumn{1}{|c|}{\text{OMA 203}} & 0.76 \priority{247}  & 0.66 \priority{30}  & 0.72 \priority{172}  &\textbf{0.81} \priority{360}  & \multicolumn{1}{c |}{0.64 \priority{0} } & 0.54 \priority{225}  & 0.38 \priority{0}  & 0.54 \priority{230}  &\textbf{0.63} \priority{360}  & \multicolumn{1}{c |}{0.44 \priority{88} }  \\ \cline{1-11}
			\multicolumn{1}{|c|}{\text{OMA 212}} & 0.80 \priority{207}  &\textbf{0.91} \priority{360}  & 0.90 \priority{349}  & 0.89 \priority{336}  & \multicolumn{1}{c |}{0.65 \priority{0} } & 0.68 \priority{296}  & 0.64 \priority{226}  & 0.49 \priority{0}  &\textbf{0.73} \priority{360}  & \multicolumn{1}{c |}{0.49 \priority{3} }  \\ \cline{1-11}
			\multicolumn{1}{|c|}{\text{OMA 281}} & 0.78 \priority{231}  & 0.66 \priority{74}  & 0.78 \priority{232}  &\textbf{0.87} \priority{360}  & \multicolumn{1}{c |}{0.60 \priority{0} } & 0.66 \priority{145}  & 0.60 \priority{0}  & 0.63 \priority{64}  &\textbf{0.76} \priority{360}  & \multicolumn{1}{c |}{0.62 \priority{52} }  \\ \cline{1-11}
			\multicolumn{1}{|c|}{\text{OMA 374}} & 0.67 \priority{236}  & 0.52 \priority{0}  & 0.75 \priority{347}  &\textbf{0.75} \priority{360}  & \multicolumn{1}{c |}{0.58 \priority{98} } & 0.60 \priority{244}  & 0.40 \priority{0}  & 0.44 \priority{38}  &\textbf{0.69} \priority{360}  & \multicolumn{1}{c |}{0.56 \priority{192} }  \\ \cline{1-11}
			\multicolumn{1}{|c|}{\text{Average}} & 0.77 \priority{227}  & 0.75 \priority{190}  & 0.78 \priority{244}  &\textbf{0.85} \priority{360}  & \multicolumn{1}{c |}{0.63 \priority{0} } & 0.55 \priority{191}  & 0.43 \priority{0}  & 0.45 \priority{39}  &\textbf{0.66} \priority{360}  & \multicolumn{1}{c |}{0.49 \priority{95} }  \\ \cline{1-11}
		\end{tabular}
  \label{Tab:Shadow_Quality}
\end{table*}

\subsection{Seasonal Specificity and Stability Tests}
\label{Sec:Season_Specific_and_Stable_Results}
Seasonal specificity refers to the capability of \ournerf{} to render seasonal features that are expected based on provided time of the year.
Seasonal stability refers to these features' invariance when changing viewing and solar angles.
The ideal method for measuring a model's ability to be specific and stable is to generate images across all reasonable viewing angles, solar angles, and time combinations.
These images could then be compared to ground truth images to show that the quality does not vary and the rendered images have the desired seasonal properties and shadows while still accurately capturing properties invariant to seasons and shadows.
While it is possible to render images for all different input configurations, we are extremely limited in the available ground truth data.
As such, we must determine an alternative method to test the specificity and stability of \ournerf{} approach.

\begin{figure*}[!tp]
	\centering
	\includegraphics[width=1.0\linewidth]{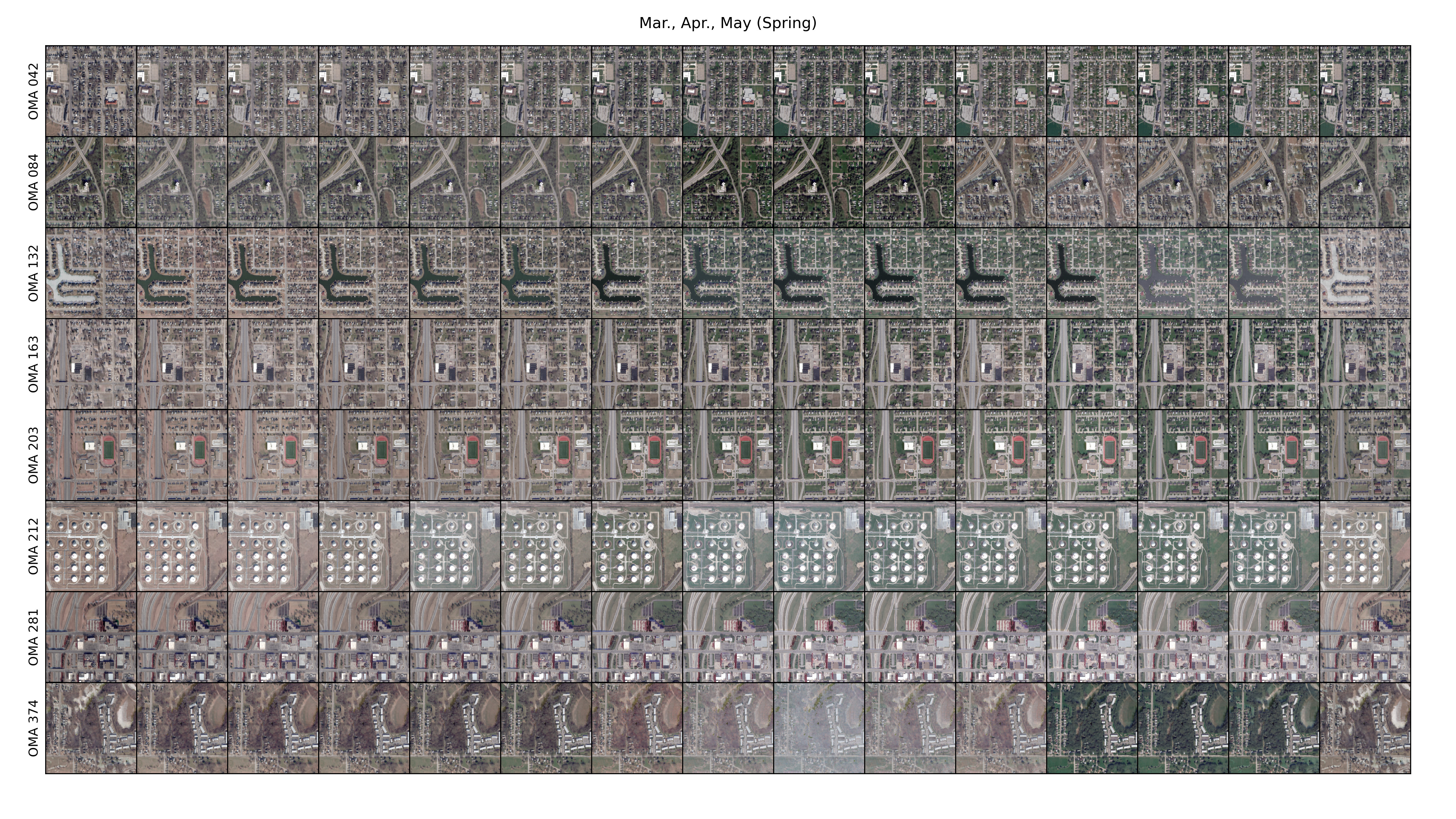}
	\caption{\label{Fig:Spring}Images rendered approximately six days apart during March, April, and May.
		They approximate the spring season.}
\end{figure*}

\begin{figure*}[!tp]
	\centering
	\includegraphics[width=1.0\linewidth]{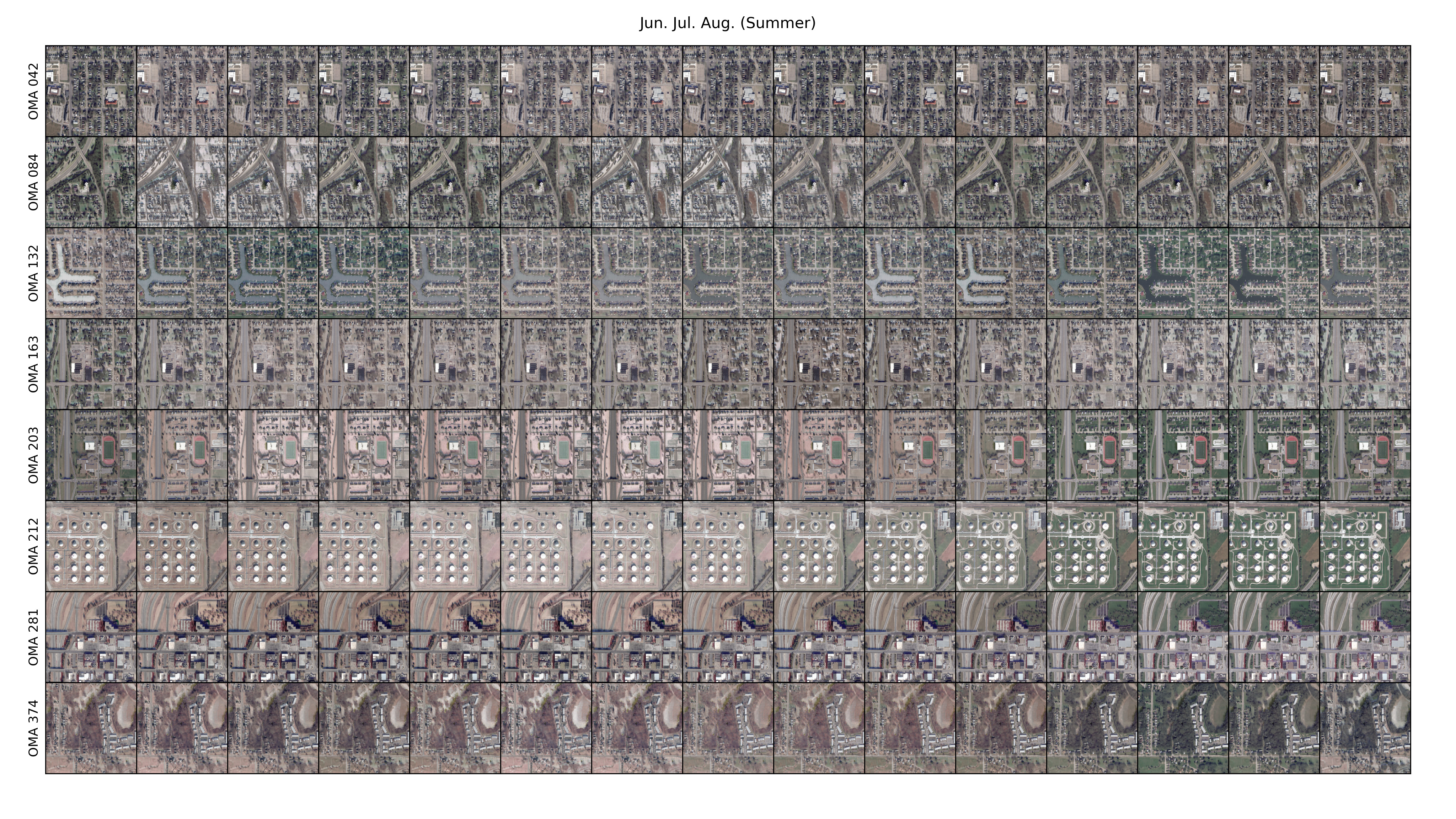}
	\caption{\label{Fig:Summer}Images rendered approximately six days apart during June, July, and August.
		They approximate the summer season.}
\end{figure*}

\begin{figure*}[!tp]
	\centering
	\includegraphics[width=1.0\linewidth]{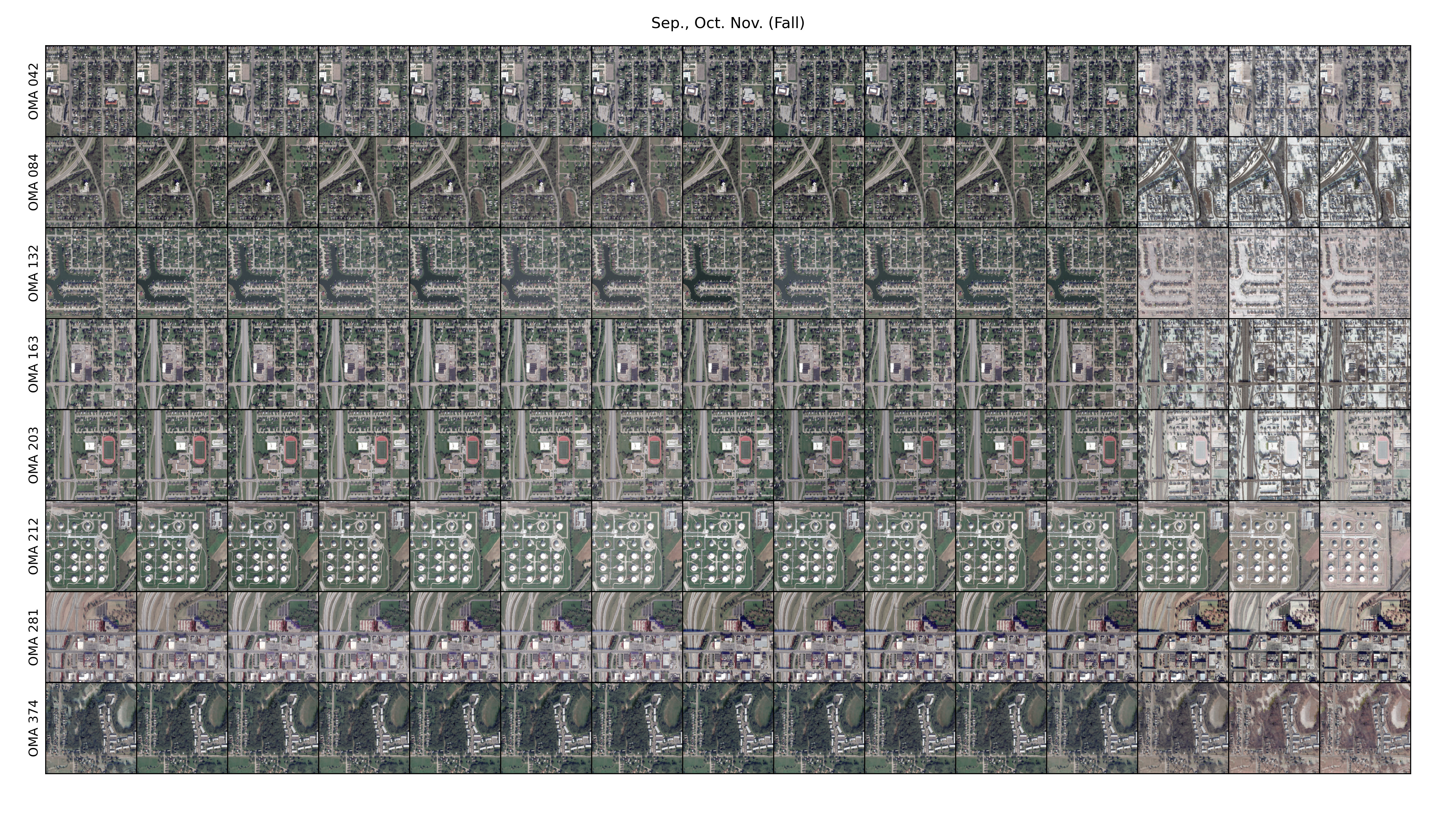}
	\caption{\label{Fig:Fall}Images rendered approximately six days apart during September, October, and November.
		They approximate the Fall season.}
\end{figure*}

\begin{figure*}[!tp]
	\centering
	\includegraphics[width=1.0\linewidth]{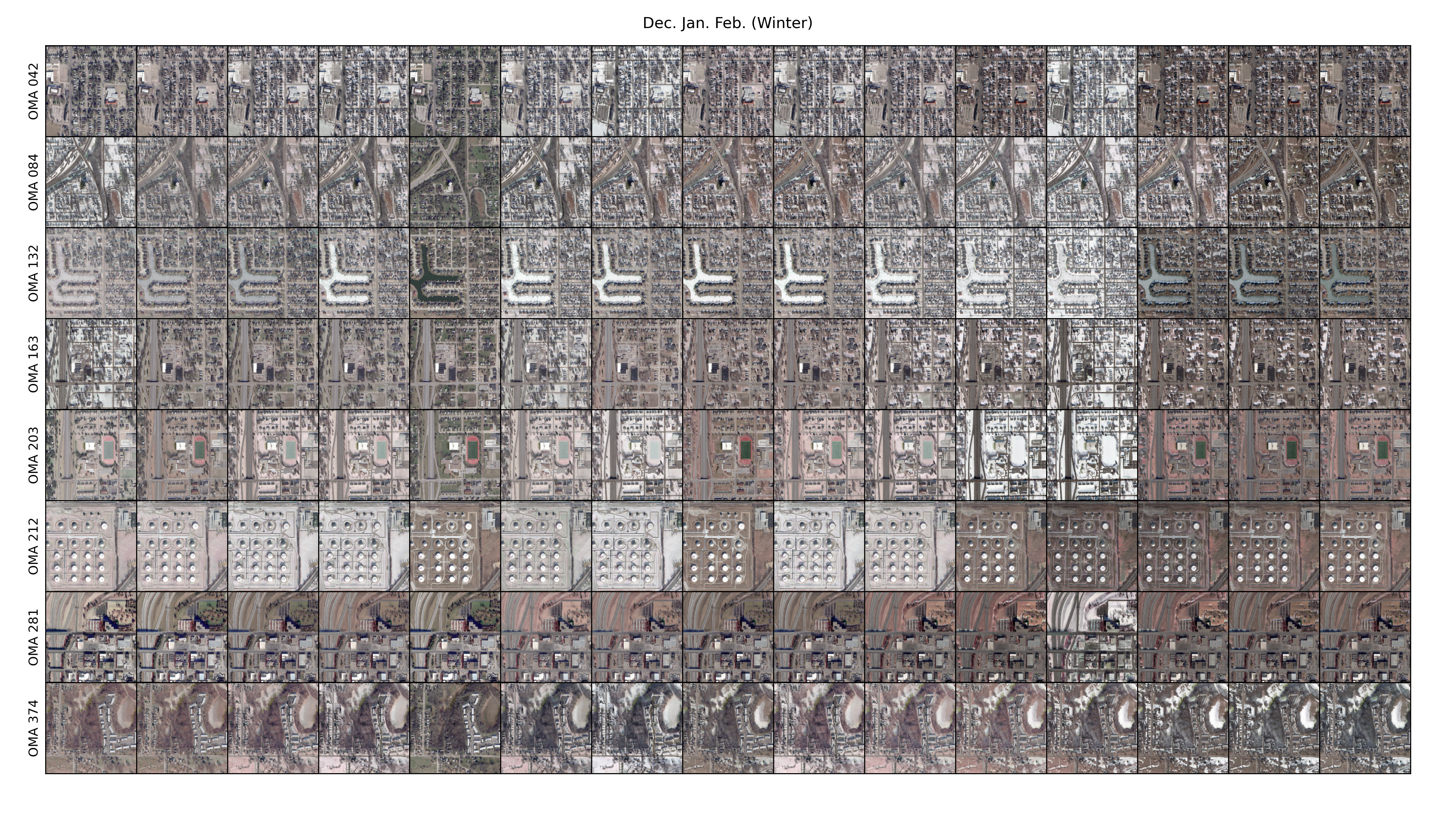}
	\caption{\label{Fig:Winter}Images rendered approximately six days apart during December, January, and February.
		They approximate the Winter season.}
\end{figure*}

To determine that \ournerf{} can render a specific season, we render 180 images, each approximately 6 days apart, spanning Spring (Fig. \ref{Fig:Spring}), Summer (Fig. \ref{Fig:Summer}), Fall (Fig. \ref{Fig:Fall}), and Winter (Fig. \ref{Fig:Winter}).
We can visually confirm the ability of \ournerf{} to render images with the expected seasonal features and correct seasonally independent properties.
The rendered images contain expected seasonal features for Spring, Fall, and Winter.
In Winter (Fig. \ref{Fig:Winter}), snow melts and returns in several regions, and in late fall (last columns of Fig. \ref{Fig:Fall}), some images have snow, and others do not, as some regions had an early snowfall and others did not.
In Spring (Fig. \ref{Fig:Spring}), we can see brown foliage becoming green; however, in the last column, certain regions seem to revert to winter.
This pattern continues in Summer (Fig. \ref{Fig:Summer}), where we see snow appear.
However, we attribute this strange behavior to the lack of data during the summer months, which explains why our model could not correctly predict the appearance of these seasons.
Despite the lack of data from summer, giving the model a time of the year during summer does not result in a distorted image.
Instead, it results in a valid image with incorrect seasonal features but correct seasonally independent features.
Thus, \ournerf{} can accurately specify the season, assuming training data is available during the desired season.

To measure the stability of the images generated by \ournerf{}, we measure the Earth Mover's Distance (EMD) \cite{EM_Dist} between images rendered at the same time of the year but with different viewing and solar angles.
Shifts in the location of structures within the image caused by changing the viewing angles should have minimal impact on the image's histogram.
Therefore, shifts in viewing angles should not significantly influence the EMD since EMD is a histogram-based image similarity metric, and we are not using a variant of EMD that considers pixel position.
As seasonal shifts generally affect the entire image, these changes should provide the bulk of the distance measured by EMD.
However, viewing and solar angle changes would still result in a non-zero EMD.

To estimate the expected EMD between two seasonally different images, we consider the EMD between the prototypical testing images.
The EMD between prototypical testing images provides a value for the expected EMD between images with different season features.
The thresholds for seasonal changes between prototypical images are summarized in Table \ref{Tab:Seasonal_Stability_Baseline}, and Table \ref{Tab:Seasonal_Stability} provides an overview of the EMD across 660 different combinations of view angle, solar angle, and time.
Each test case is computed at one of five solar angles, eleven viewing angles, and twelve viewing times.
These points are shown in Fig. \ref{Fig:Data_Example}.
We are primarily interested in Cases A and B, given that Cases C, D, and E have inferior performance in other areas.
However, we include all cases in Table \ref{Tab:Seasonal_Stability} for completeness.

\begin{figure}[!tp]
	\centering
	\includegraphics[width=1.0\linewidth]{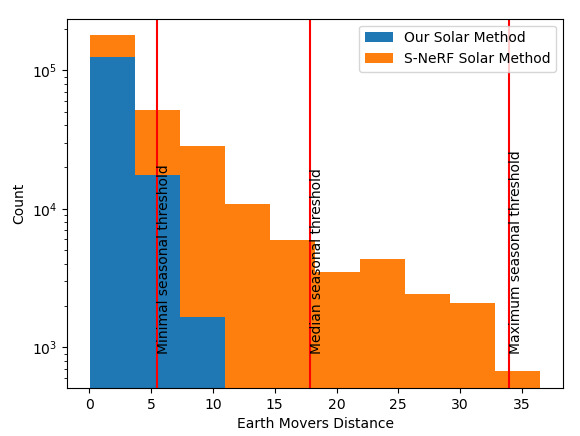}
	\caption{\label{Fig:Seasonal_Stab_Hist}A histogram of the EMD amongst images with varying views and solar angles but a fixed time of the year.}
\end{figure}

\begin{figure}[!tp]
	\centering
	\includegraphics[width=1.0\linewidth]{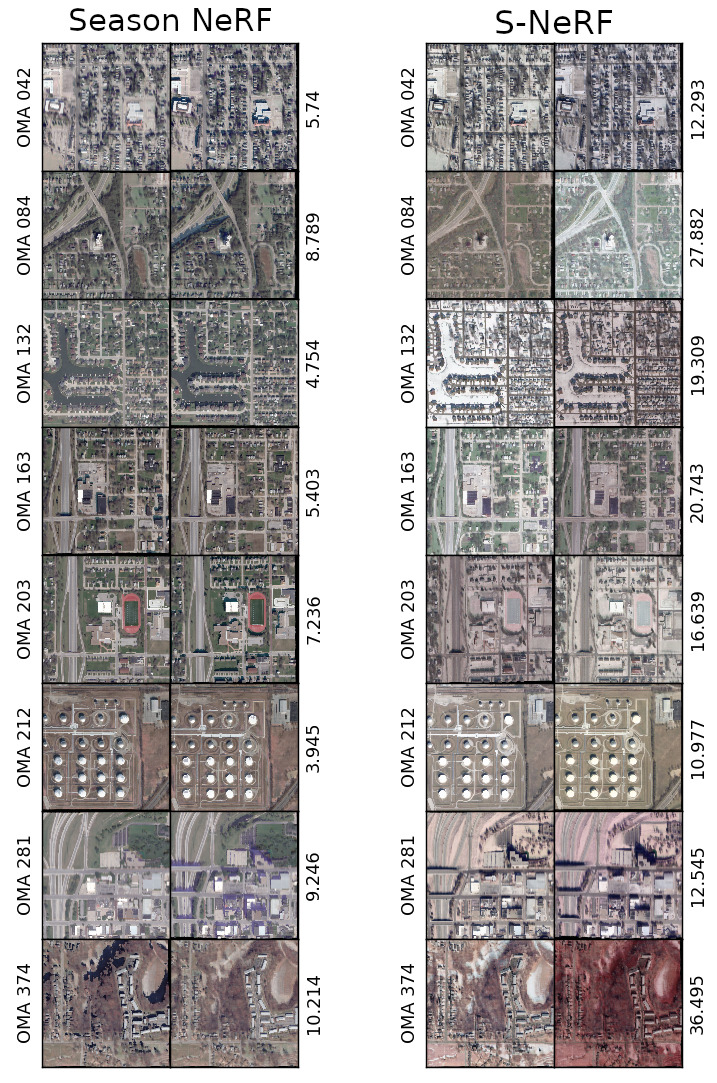}
	\caption{\label{Fig:Seasonal_Stab_Examples}Examples of the image pairs with the largest EMD for each region.
		The EM distance between the two images is on the right side of each column.}
\end{figure}

We compare the results of our method for rendering shadows with those of S-NeRF's method of rendering shadows in Figs. \ref{Fig:Seasonal_Stab_Examples} and  \ref{Fig:Seasonal_Stab_Hist}.
In Fig. \ref{Fig:Seasonal_Stab_Examples}, we show the image pairs from each region with the largest EMD when the time of the year is constant.
We provide a histogram of the EMD of all the image pairs we tested in Fig. \ref{Fig:Seasonal_Stab_Hist}.
Based on the histogram in Fig. \ref{Fig:Seasonal_Stab_Hist} and rendered images in Fig. \ref{Fig:Seasonal_Stab_Examples}, we conclude our shadow rendering process results in a seasonally stable network, where solar and view angle do not alter seasonal features.
Using the solar approach from S-NeRF does not ensure that the seasonal features are independent of the solar angle.
We conclude this because a significant portion of the image pairs rendered with S-NeRF's shadow method had an EMD above the threshold set by the prototypical images, indicating a change in the seasonal features.

\begin{table}[!htb]
	\centering
	\caption{\label{Tab:Seasonal_Stability_Baseline}Baseline EMD for seasonal effects}
	\begin{tabular}{|c|S[table-format=2.2, round-precision= 2, round-mode=places] | S[table-format=2.2, round-precision= 2, round-mode=places] | S[table-format=2.2, round-precision= 2, round-mode=places] | }
		\hline
		\text{Baseline Scores} & \multicolumn{1}{c|}{\text{Min}} & \multicolumn{1}{c|}{\text{Median}} & \multicolumn{1}{c|}{\text{Max}} \\ \hline
		\text{OMA 042} & 13.234352111816406 & 16.297462463378906 & 19.17926597595215  \\ \hline
		\text{OMA 084} & 8.756423950195312 & 23.066667556762695 & 23.24638557434082  \\ \hline
		\text{OMA 132} & 9.183882713317871 & 18.830217361450195 & 23.95235824584961  \\ \hline
		\text{OMA 163} & 9.649008750915527 & 16.639848709106445 & 19.758256912231445  \\ \hline
		\text{OMA 203} & 6.006046295166016 & 23.42717170715332 & 24.46904754638672  \\ \hline
		\text{OMA 212} & 17.878915786743164 & 25.50100326538086 & 33.948997497558594  \\ \hline
		\text{OMA 281} & 11.18840503692627 & 11.387837409973145 & 12.050569534301758  \\ \hline
		\text{OMA 374} & 5.476902008056641 & 17.834917068481445 & 18.008724212646484  \\ \hline
		\text{Average} & 10.17174208164215 & 19.123140692710876 & 21.826700687408447  \\ \hline
	\end{tabular}
\end{table}

\begin{table}[!htb]
	\centering
	\caption{\label{Tab:Seasonal_Stability}Quantitative overview of EMD when time is constant. Case A: Full Model.  Case B: Full Model, S-NeRF Solar loss.  Case C: MSE Loss.  Case D: No Height Map.  Case E: No Seasonal Adjustment.}
	\begin{tabular}{c c c c c c }
		\hline
		\multicolumn{1}{|c|}{\text{Reg}} & \multicolumn{5}{c|}{\text{Median $\downarrow$}} \\ \hline
		\multicolumn{1}{|c|}{\text{}}& \multicolumn{1}{c}{\text{A}} & \multicolumn{1}{c}{\text{B}} & \multicolumn{1}{c}{\text{C}} & \multicolumn{1}{c}{\text{D}} & \multicolumn{1}{c|}{\text{E}} \\ \hline
		\multicolumn{1}{|c|}{\text{042}} & 1.60 \priority{346}  & 3.21 \priority{0}  & 1.81 \priority{301}  &\textbf{1.54} \priority{360}  & \multicolumn{1}{c |}{2.28 \priority{201} }  \\ \hline
		\multicolumn{1}{|c|}{\text{084}} & 2.24 \priority{333}  & 12.16 \priority{0}  & 1.78 \priority{348}  &\textbf{1.44} \priority{360}  & \multicolumn{1}{c |}{3.01 \priority{307} }  \\ \hline
		\multicolumn{1}{|c|}{\text{132}} & 1.49 \priority{351}  & 7.40 \priority{0}  & 1.76 \priority{335}  &\textbf{1.35} \priority{360}  & \multicolumn{1}{c |}{1.76 \priority{335} }  \\ \hline
		\multicolumn{1}{|c|}{\text{163}} & 1.76 \priority{342}  & 8.28 \priority{0}  & 1.47 \priority{357}  &\textbf{1.43} \priority{360}  & \multicolumn{1}{c |}{2.20 \priority{319} }  \\ \hline
		\multicolumn{1}{|c|}{\text{203}} & 2.46 \priority{311}  & 6.28 \priority{0}  & 2.35 \priority{319}  &\textbf{1.86} \priority{360}  & \multicolumn{1}{c |}{2.52 \priority{306} }  \\ \hline
		\multicolumn{1}{|c|}{\text{212}} & 1.43 \priority{169}  & 1.51 \priority{28}  &\textbf{1.33} \priority{360}  & 1.52 \priority{0}  & \multicolumn{1}{c |}{1.52 \priority{0} }  \\ \hline
		\multicolumn{1}{|c|}{\text{281}} & 2.42 \priority{346}  & 6.86 \priority{0}  & 2.56 \priority{335}  &\textbf{2.24} \priority{360}  & \multicolumn{1}{c |}{4.94 \priority{149} }  \\ \hline
		\multicolumn{1}{|c|}{\text{374}} & 2.45 \priority{231}  & 4.59 \priority{0}  & 1.77 \priority{303}  &\textbf{1.25} \priority{360}  & \multicolumn{1}{c |}{2.18 \priority{260} }  \\ \hline
		\multicolumn{1}{|c|}{\text{Avg}} & 1.98 \priority{329}  & 6.29 \priority{0}  & 1.85 \priority{339}  &\textbf{1.58} \priority{360}  & \multicolumn{1}{c |}{2.55 \priority{285} }  \\ \hline
		& & & & &\\
		\hline
		\multicolumn{1}{|c|}{\text{Reg}} & \multicolumn{5}{c|}{\text{95\% Quantile $\downarrow$}} \\ \hline
		\multicolumn{1}{|c|}{\text{}}& \multicolumn{1}{c}{\text{A}} & \multicolumn{1}{c}{\text{B}} & \multicolumn{1}{c}{\text{C}} & \multicolumn{1}{c}{\text{D}} & \multicolumn{1}{c|}{\text{E}} \\ \hline
		\multicolumn{1}{|c|}{\text{042}} & 3.57 \priority{328}  & 8.55 \priority{0}  & 3.99 \priority{300}  &\textbf{3.08} \priority{360}  & \multicolumn{1}{c |}{4.72 \priority{251} }  \\ \hline
		\multicolumn{1}{|c|}{\text{084}} & 5.39 \priority{314}  & 24.37 \priority{0}  & 3.85 \priority{340}  &\textbf{2.66} \priority{360}  & \multicolumn{1}{c |}{5.41 \priority{314} }  \\ \hline
		\multicolumn{1}{|c|}{\text{132}} & 3.07 \priority{342}  & 15.17 \priority{0}  & 3.91 \priority{318}  &\textbf{2.46} \priority{360}  & \multicolumn{1}{c |}{3.32 \priority{335} }  \\ \hline
		\multicolumn{1}{|c|}{\text{163}} & 3.38 \priority{347}  & 17.31 \priority{0}  & 3.02 \priority{356}  &\textbf{2.89} \priority{360}  & \multicolumn{1}{c |}{4.16 \priority{328} }  \\ \hline
		\multicolumn{1}{|c|}{\text{203}} & 4.82 \priority{310}  & 13.29 \priority{0}  & 5.17 \priority{297}  &\textbf{3.46} \priority{360}  & \multicolumn{1}{c |}{4.91 \priority{306} }  \\ \hline
		\multicolumn{1}{|c|}{\text{212}} & 2.54 \priority{353}  & 8.91 \priority{0}  &\textbf{2.43} \priority{360}  & 2.72 \priority{344}  & \multicolumn{1}{c |}{2.89 \priority{334} }  \\ \hline
		\multicolumn{1}{|c|}{\text{281}} & 6.03 \priority{296}  & 9.69 \priority{0}  & 7.14 \priority{206}  &\textbf{5.25} \priority{360}  & \multicolumn{1}{c |}{8.14 \priority{125} }  \\ \hline
		\multicolumn{1}{|c|}{\text{374}} & 7.69 \priority{293}  & 32.26 \priority{0}  & 4.77 \priority{328}  &\textbf{2.17} \priority{360}  & \multicolumn{1}{c |}{7.64 \priority{294} }  \\ \hline
		\multicolumn{1}{|c|}{\text{Avg}} & 4.56 \priority{319}  & 16.19 \priority{0}  & 4.29 \priority{327}  &\textbf{3.09} \priority{360}  & \multicolumn{1}{c |}{5.15 \priority{303} }  \\ \hline
		& & & & &\\
		\cline{1-6}
		\multicolumn{1}{|c|}{\text{Reg}} & \multicolumn{5}{c|}{\text{Max $\downarrow$}} \\ \cline{1-6}
		\multicolumn{1}{|c|}{\text{}}& \multicolumn{1}{c}{\text{A}} & \multicolumn{1}{c}{\text{B}} & \multicolumn{1}{c}{\text{C}} & \multicolumn{1}{c}{\text{D}} & \multicolumn{1}{c|}{\text{E}} \\ \cline{1-6}
		\multicolumn{1}{|c|}{\text{042}} & 5.74 \priority{337}  & 12.29 \priority{0}  & 6.53 \priority{297}  &\textbf{5.31} \priority{360}  & \multicolumn{1}{c |}{6.49 \priority{299} }  \\ \cline{1-6}
		\multicolumn{1}{|c|}{\text{084}} & 8.79 \priority{279}  & 27.88 \priority{0}  & 6.68 \priority{310}  &\textbf{3.33} \priority{360}  & \multicolumn{1}{c |}{7.52 \priority{298} }  \\ \cline{1-6}
		\multicolumn{1}{|c|}{\text{132}} & 4.75 \priority{343}  & 19.31 \priority{0}  & 6.41 \priority{304}  &\textbf{4.04} \priority{360}  & \multicolumn{1}{c |}{4.27 \priority{354} }  \\ \cline{1-6}
		\multicolumn{1}{|c|}{\text{163}} & 5.40 \priority{330}  & 20.74 \priority{0}  & 4.59 \priority{347}  &\textbf{4.02} \priority{360}  & \multicolumn{1}{c |}{5.50 \priority{328} }  \\ \cline{1-6}
		\multicolumn{1}{|c|}{\text{203}} & 7.24 \priority{277}  & 16.64 \priority{0}  & 7.82 \priority{260}  &\textbf{4.45} \priority{360}  & \multicolumn{1}{c |}{6.57 \priority{297} }  \\ \cline{1-6}
		\multicolumn{1}{|c|}{\text{212}} & 3.94 \priority{352}  & 10.98 \priority{0}  & 4.44 \priority{327}  &\textbf{3.80} \priority{360}  & \multicolumn{1}{c |}{5.41 \priority{279} }  \\ \cline{1-6}
		\multicolumn{1}{|c|}{\text{281}} & 9.25 \priority{309}  & 12.55 \priority{0}  & 10.34 \priority{206}  &\textbf{8.71} \priority{360}  & \multicolumn{1}{c |}{10.24 \priority{216} }  \\ \cline{1-6}
		\multicolumn{1}{|c|}{\text{374}} & 10.21 \priority{318}  & 36.49 \priority{0}  &\textbf{6.83} \priority{360}  & 7.10 \priority{356}  & \multicolumn{1}{c |}{9.04 \priority{333} }  \\ \cline{1-6}
		\multicolumn{1}{|c|}{\text{Avg}} & 6.92 \priority{314}  & 19.61 \priority{0}  & 6.71 \priority{320}  &\textbf{5.10} \priority{360}  & \multicolumn{1}{c |}{6.88 \priority{315} }  \\ \cline{1-6}
	\end{tabular}
\end{table}
\subsection{Analysis of Barron's Loss Compared to MSE Loss}
\label{Sec:Barron_MSE_Compare}
%A simple comparison of visual quality scores in Table \ref{Tab:Image_Quality} suggests Barron's Loss does not provide a noticeable improvement over MSE loss.
%However, these scores do not provide a complete picture of performance as they compare rendered images to images containing transient features.
%As a result, removing transient objects in a rendered image does not necessarily improve the visual similarity scores, as the ground truth image contains transient objects.
%Furthermore, as noted in Derksen and Izzo \cite{SNeRF}, including shadow prediction capability tends to use shadows to darken areas with transient objects.
%Examples of this occurring when \ournerf{} uses MSE loss are shown in Fig. FIGURE.
%Also shown in Fig. FIGURE are images using Barron's Loss instead.
%From those examples, we conclude that Barron's Loss removes %tansient objects and provides a reasonable estimate for the color of regions occluded by transient objects in the training data.

A comparison of visual quality scores in Table \ref{Tab:Image_Quality} and height map quality in Table \ref{Tab:HM_Quality} models trained using Barron's loss perform almost identically to those trained with MSE loss.
However, the models have different performances in shadow prediction, with Barron's loss improving shadow recall by an average of 13\%.
The change in performance is unexpected, as shadow prediction never uses Barron's loss.
Determining an explanation for this phenomenon is an area for future work.

\subsection{High Resolution Results}
\begin{figure}[!tp]
	   \centering
	   \includegraphics[width=1.0\linewidth]{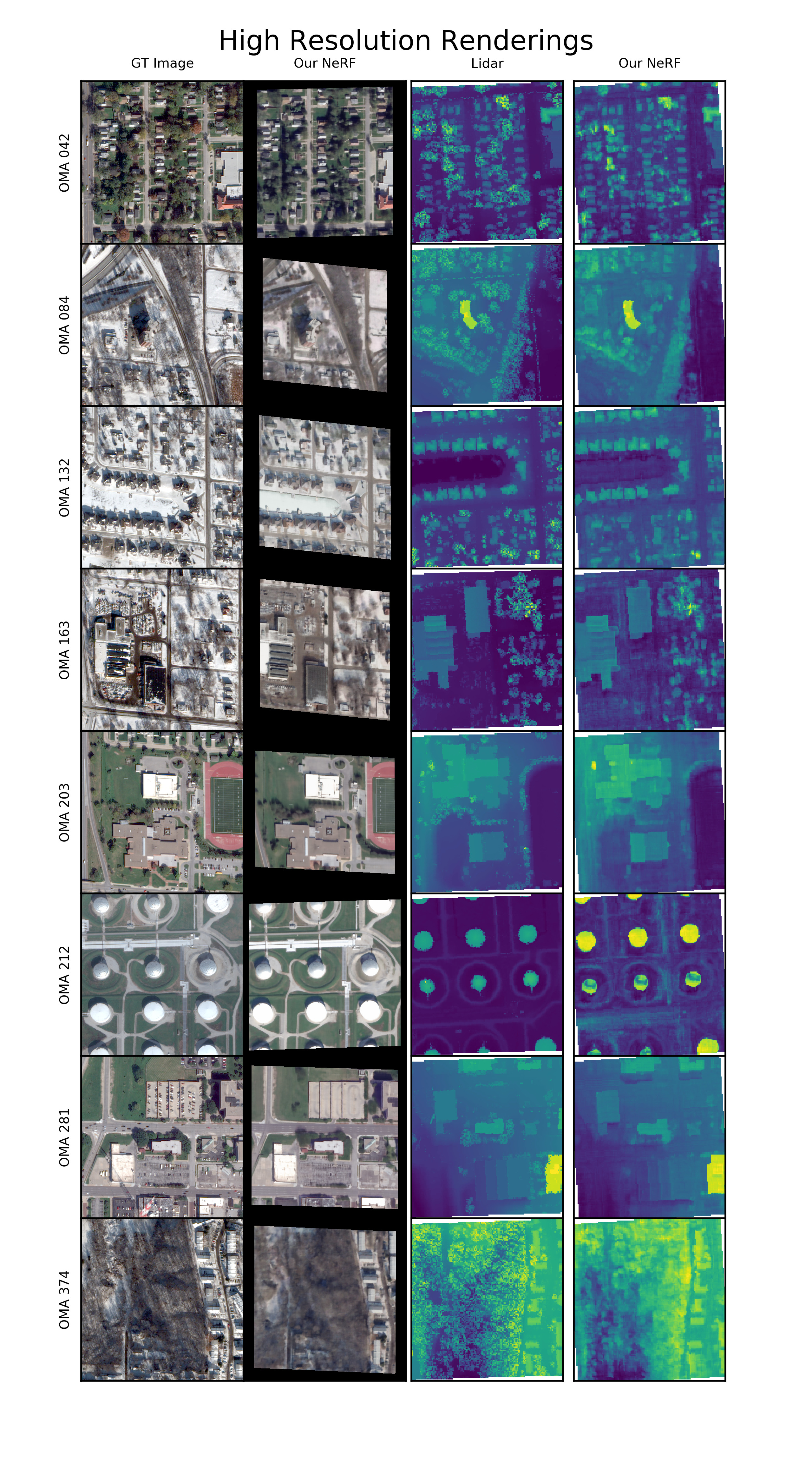}
	   \caption{\label{Fig:Rend_HR}Results with high resolution images on small regions.}
    \end{figure}

	\textcolor{Changes}{We also consider how increasing the resolution of the images used to create \ournerf{} affects the performance.
			To accomplish this, we do not down-sample the images; however, we must reduce the size of the regions we consider from 500 by 500 meters to 250 by 250 meters.
			We consider using parameters tuned for large area low-resolution images and parameters tuned for small area high resolution.
			The results are shown in Table \ref{Tab:Small_Regions}, with example renderings in Fig. \ref{Fig:Rend_HR}.
			Unsurprisingly, tuning the parameters results in superior performance.
			We show the changed parameters in Table \ref{Tab_param_tuning}.}
		
		\begin{table*}[!htb]
			\centering
			\caption{\label{Tab:Small_Regions} Results on small regions. Case A: Full Model on Full Region. Case F: Full Model without downsizing and parameters tuned for small regions.  Case G: Full model without down-sampling default parameters.}
			\begin{tabular}{|c|c c c | c c c | }
				\hline
				\text{Region} & \multicolumn{3}{c|}{\text{SSIM, SA $\uparrow$}} & \multicolumn{3}{c|}{\text{MAE $\downarrow$}} \\ \hline
				\text{Case:} & \text{A} & \text{F} & \text{G} & \text{A} & \text{F} & \text{G} \\ \hline
				\text{OMA 042} &\textbf{0.602} \priority{360} &  0.586 \priority{218} &  0.562 \priority{0} &  2.473 \priority{0} &\textbf{1.957} \priority{360} &  2.357 \priority{80}  \\ \hline
				\text{OMA 084} &\textbf{0.571} \priority{360} &  0.566 \priority{310} &  0.529 \priority{0} &  2.823 \priority{256} &\textbf{2.442} \priority{360} &  3.769 \priority{0}  \\ \hline
				\text{OMA 132} &  0.674 \priority{0} &  0.702 \priority{333} &\textbf{0.704} \priority{360} &  1.219 \priority{89} &\textbf{1.040} \priority{360} &  1.279 \priority{0}  \\ \hline
				\text{OMA 163} &\textbf{0.585} \priority{360} &  0.532 \priority{25} &  0.527 \priority{0} &\textbf{1.431} \priority{360} &  1.466 \priority{340} &  2.073 \priority{0}  \\ \hline
				\text{OMA 203} &  0.656 \priority{0} &\textbf{0.705} \priority{360} &  0.696 \priority{293} &  1.171 \priority{209} &\textbf{0.987} \priority{360} &  1.429 \priority{0}  \\ \hline
				\text{OMA 212} &  0.610 \priority{0} &\textbf{0.711} \priority{360} &  0.679 \priority{245} &  0.939 \priority{274} &\textbf{0.889} \priority{360} &  1.102 \priority{0}  \\ \hline
				\text{OMA 281} &\textbf{0.636} \priority{360} &  0.634 \priority{322} &  0.618 \priority{0} &  1.461 \priority{305} &\textbf{1.312} \priority{360} &  2.296 \priority{0}  \\ \hline
				\text{OMA 374} &\textbf{0.533} \priority{360} &  0.464 \priority{0} &  0.473 \priority{48} &  4.436 \priority{0} &  4.216 \priority{351} &\textbf{4.211} \priority{360}  \\ \hline
				\text{Average} &  0.608 \priority{254} &\textbf{0.613} \priority{360} &  0.599 \priority{0} &  1.994 \priority{219} &\textbf{1.789} \priority{360} &  2.314 \priority{0}  \\ \hline
			\end{tabular}
		\end{table*}

		\begin{table*}[!htb]
			\centering
			\caption{\label{Tab_param_tuning}
				A summary of the initial and best hyperparameters found as a result of the hyperparamter tuning process.
				$LR$: learning rate, $\lambda_{SC}$ weight for solar correction terms, $\lambda_{DS}$ weight for depth supervision terms, n: number of seasonal classes $\kappa$: scale for shadow mask $\mu$: Shift for shadow mask
				$\mathbb{S}$: Maximum sky color
			$\mathbb{A}$: Minimum color without shadows}
			\begin{tabular}{|c|c | c | c | c | c | c | c | c | c | c | }
				\hline
				\text{Region} & \multicolumn{1}{c|}{\text{$\log_{10}\left(LR\right)$}} & \multicolumn{1}{c|}{\text{$\lambda_{SC}$}} & \multicolumn{1}{c|}{\text{$\lambda_{DS}$}} & \multicolumn{1}{c|}{\text{n}} & \multicolumn{1}{c|}{\text{$\kappa$}} & \multicolumn{1}{c|}{\text{$\mu$}} & \multicolumn{1}{c|}{\text{$\mathbb{S}$}} & \multicolumn{1}{c|}{\text{$\mathbb{A}$}} & \multicolumn{1}{c|}{\text{SSIM}} & \multicolumn{1}{c|}{\text{MAE}} \\ \hline
				\text{Low Res} & -4.84 & 0.03 & 1.00 & 4 & 30.00 & 0.20 & 0.50 & 0.20 & 0.53 & 2.52   \\ \hline
				\text{High Res} & -4.37 & 0.46 & 2.63 & 8 & 37.91 & 0.10 & 0.73 & 0.15 & 0.57 & 1.70 \\ \hline
			\end{tabular}
		\end{table*}

\section{Conclusion}
\label{Sec:Conclusion}
\begin{table}
\centering
\caption{\label{Tab_Summary} Summary of the performances for each of the variations of \ournerf{}.
Case A: Full Model.  Case B: Full Model, S-NeRF Solar loss.  Case C: MSE Loss.  Case D: No Height Map.  Case E: No Seasonal Adjustment.}
\begin{tblr}{
  cells = {c},
  cell{2}{2} = {green},
  cell{2}{3} = {green},
  cell{2}{4} = {green},
  cell{2}{5} = {green},
  cell{3}{2} = {green},
  cell{3}{4} = {green},
  cell{3}{5} = {green},
  cell{3}{6} = {green},
  cell{4}{2} = {green},
  cell{4}{3} = {green},
  cell{4}{4} = {green},
  cell{5}{2} = {green},
  cell{5}{5} = {green},
  hlines,
  vlines,
}
Cases:             & A & B & C & D & E \\
Seasonal Variation & X & X & X & X &   \\
Seasonal Stability & X &   & X & X & X \\
Accurate DSM       & X & X & X &   &   \\
Good Shadow Recall & X &   &   & X &  
\end{tblr}
\end{table}

Given multidate and multiview satellite images of a scene, we can render novel views with specific solar and seasonal features using a NeRF.
Since seasonal features are obviously linked to the time of the year, \ournerf{} uses the time of the year to compute seasonally adjusted albedo colors.
In addition, \ournerf{} uses the viewing and solar angles to determine which regions of the rendered image contain shadows.
The network architecture of \ournerf{} ensures the time of the year can only alter the seasonal features by limiting the time of the year to change the seasonally adjusted albedo and leaving other outputs independent of the time of the year.
While the method for rendering shadows described in S-NeRF \cite{SNeRF} works well, it allows the solar angle to influence the seasonal features unduly.
To discourage the network from using the seasonally adjusted albedo to explain shadows, we alter how shadows are computed and modify the loss function.
We expand the loss function to punish the network for outputting dark colors via the seasonally adjusted albedo.
Furthermore, we punish the network for using trivial sky colors.
Finally, we freeze parameters not used in the computation of the solar visibility or sky color when training the network with solar rays and freeze parameters used only in the computation of solar visibility when training with image rays.
The ability to render seasonal features can also cause the network to learn less accurate densities.
To balance this tendency, we provide a height map to our model as a guide for the early stages of training.

The primary purpose of \ournerf{} is to account for seasonal changes within satellite images.
To this end, we introduced temporal adjustment capability into our NeRF process.
In addition to this, \ournerf{} uses a novel approach for depth supervision that utilizes a height map rather than a set of 3D points as is used on \cite{Depth_Supervised_NeRF}.
\textcolor{Changes}{In the future, we would like to compare the performance of NeRFs using different depth supervision techniques with varying quality of prior information.}
\textcolor{Changes}{In addition, we would like to incorporate the findings of \cite{Robust_NeRF} to improve the performance of transient object removal for \ournerf{}.}

We show the performance of \ournerf{} in eight AOIs from the \textit{2019 IEEE GRSS Data Fusion Contest}
\cite{Sat_data_2}, which contains images captured by the Maxar
WorldView-3 satellite between 2014 and 2016.
We show how, by including seasonal variation, we can improve the quality of the rendered image as measured by PSNR and SSIM.
By including a height map, we can improve the mean altitude error over every region by an average of .6 meters.
Using Barron's loss function, we can improve the average shadow recall by 13\%.  However, Barron's loss does not have a discernible impact on the quality of the height map or rendered image.
\textcolor{Changes}{As summarized in Table \ref{Tab_Summary}, each aspect of the \ournerf{} framework contributes to the capabilities of the radiance field.}
Our method for computing shadows allows solar features to remain independent from seasonal features despite a limited amount of training data.
As a result, we can specify solar and seasonal features for novel view renderings of scenes captured by satellite images.

%%%%%%%%% REFERENCES
{
	\bibliographystyle{IEEEtran}
	\bibliography{bibliography}
}

% \includegraphics command within an optional argument.  (You could create

\begin{IEEEbiography}[{\includegraphics[width=1in,height=1.25in,clip,keepaspectratio]{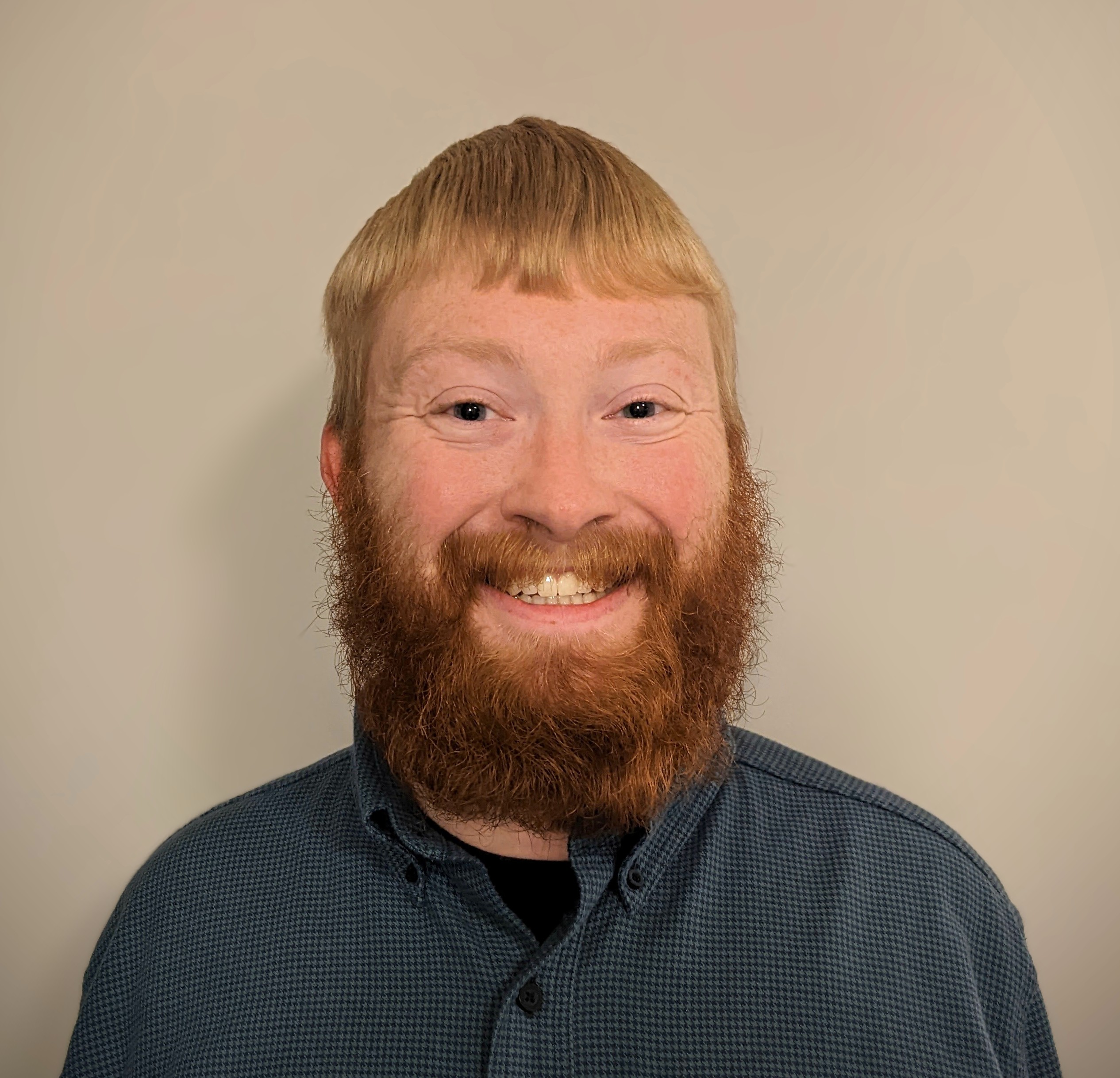}}]{Michael Gableman}
	is a graduate student in the School of Electrical and Computer Engineering School at Purdue University, West Lafayette, IN, USA.  He received a bachelor's degree in mathematics and computer science from Ripon College, Ripon, WI, USA, in 2016 and a master's degree in computer and engineering from the School of Electrical and Computer Engineering, Purdue University, West Lafayette, IN, USA in 2019.  His research interests include computer vision, machine learning, neural radiance fields, and satellite imagery.
\end{IEEEbiography}

%\newpage

\begin{IEEEbiography}[{\includegraphics[width=1in,height=1.25in,clip,keepaspectratio]{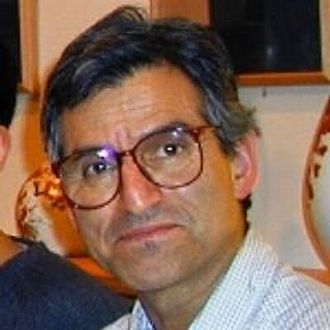}}]{Avinash C. Kak}
	received the Ph.D. degree in electrical engineering from the Indian Institute of Technology, Delhi, India, in 1970.
	He is currently a Professor of electrical and
	computer engineering with Purdue University, West
	Lafayette, IN, USA. His coauthored book Principles of Computerized Tomographic Imaging (SIAM,
	2001) was republished as a classic in applied mathematics. His other coauthored book, Digital Picture
	Processing, also considered by many to be a classic in
	computer vision and image processing, was published
	by Academic Press in 1982. His more recent books were written for his “Objects
	Trilogy” project. All three books of the trilogy have been published by John
	Wiley \& Sons. The first, Programming with Objects, came out in 2003, the
	second, Scripting with Objects, in 2008, and the last, Designing with Objects, in
	2015. His research interests include algorithms, languages, and systems related
	to wired and wireless camera networks, robotics, computer vision, etc.
\end{IEEEbiography}
\end{document}